\ifdefined\pdfminorversion\pdfminorversion=7\relax\fi
\PassOptionsToPackage{section}{placeins}
\documentclass{dotsstudio}
\usepackage{amsmath,amssymb}
\usepackage{algorithm}
\usepackage{algpseudocode}
\usepackage{longtable}
\usepackage{needspace}
\hypersetup{pdftitle={On the Design of Hyper-Connections: Simplification, Stability, and Scalability},pdfauthor={Xiangdong Zhang, Xiaohan Qin, Sunan Zou, Tuo Dai, Bizhao Shi, Zhuo Xia, Yebin Yang, Yizhou Chen, Yuliang Liu, Yu Cheng, Junchi Yan}}

\title{On the Design of Hyper-Connections: Simplification, Stability, and Scalability}
\author[1,2]{Xiangdong Zhang}
\author[1,2]{Xiaohan Qin}
\author[2,3]{Sunan Zou}
\author[2]{Tuo Dai}
\author[2]{Bizhao Shi}
\author[1,2]{Zhuo Xia}
\author[1,2]{Yebin Yang}
\author[2]{Yizhou Chen}
\author[4]{Yuliang Liu}
\author[5]{Yu Cheng}
\author[1,\dagger]{Junchi Yan}
\affiliation[1]{School of AI, Shanghai Jiao Tong University\quad
  \textsuperscript{2}Dots Studio, Xiaohongshu Inc.}
\affiliation[3]{School of CS, Peking University\quad
  \textsuperscript{4}Independent Researcher}
\affiliation[5]{Nanyang Technological University}
\contribution[\dagger]{Corresponding author: \texttt{yanjunchi@sjtu.edu.cn}.}
\code{https://github.com/aHapBean/SimpleHC}
\emailresource{zhangxiangdong@sjtu.edu.cn}

\abstract{
Hyper-Connections (HC) architectures are increasingly adopted in frontier
language models, yet their components remain insufficiently understood
and introduce non-negligible training overhead.
In this work, we systematically study the HC architecture to understand
its components and identify simpler, more efficient designs.
We find that learned read/write gating without mixing and unconstrained
mixing with fixed gates form two effective interaction groups:
either alone achieves performance comparable to mHC.
We also identify repeated access to the expanded residual state
as a major source of training overhead, even without mixing.
These findings lead to \textbf{SimpleHC}, which simplifies the architecture
by removing residual mixing and simplifies execution through
\emph{shared memory access}, using the resulting additive structure
and shared gate inputs.
Layer-specific gates share a group-entry state, enabling batched reads
and deferred writes; scalar corrections exactly preserve preceding sublayer
output contributions.
Our analyses support both the retained residual structure and the
design of shared memory access.
SimpleHC incurs only $\mathbf{2.4\%}$ extra end-to-end training time
over a vanilla residual baseline at the largest tested scale.
It maintains performance comparable to mHC when scaling from 15B to
30B MoE models, exhibits smoother gradient dynamics during standard
training, and remains stable under stress.
We hope SimpleHC will encourage a broader pursuit of simplicity in foundation
model design, making performance gains more accessible and practical at scale.
}

\begin{document}
\maketitle

\section{Introduction}

\label{sec:intro}

\begin{figure}[!b]

\centering
\begin{minipage}[t]{0.32\textwidth}
  \centering
  \includegraphics[width=\linewidth]{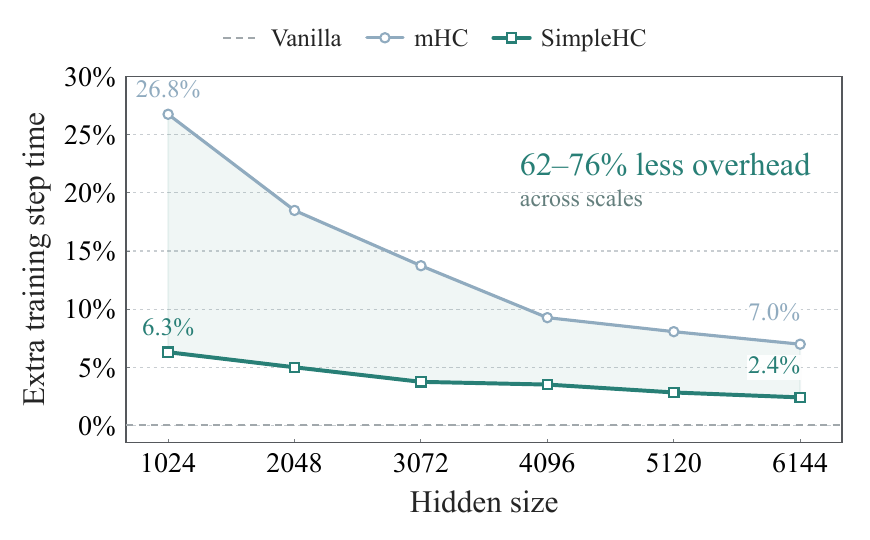}
  \par
  {\small (a) Training overhead}
\end{minipage}\hfill
\begin{minipage}[t]{0.32\textwidth}
  \centering
  \includegraphics[width=\linewidth]{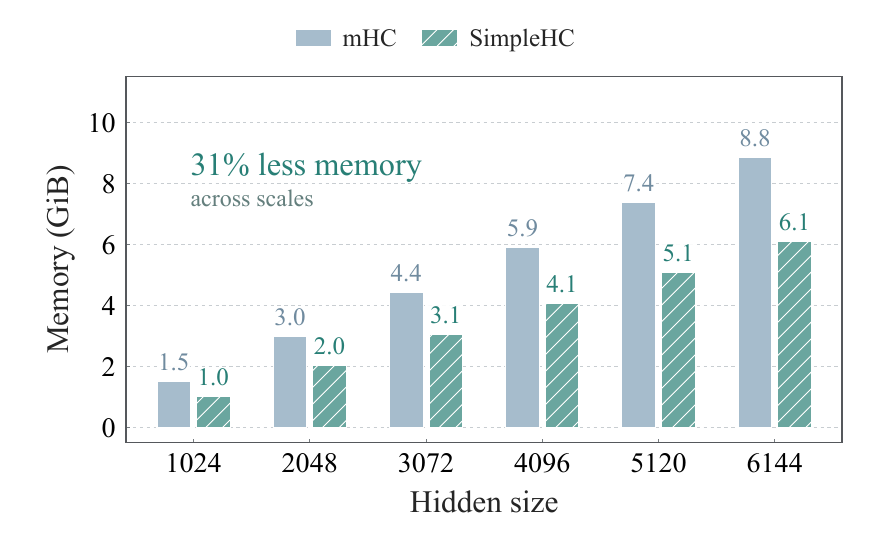}
  \par
  {\small (b) Saved activation storage}
\end{minipage}\hfill
\begin{minipage}[t]{0.32\textwidth}
  \centering
  \includegraphics[width=\linewidth]{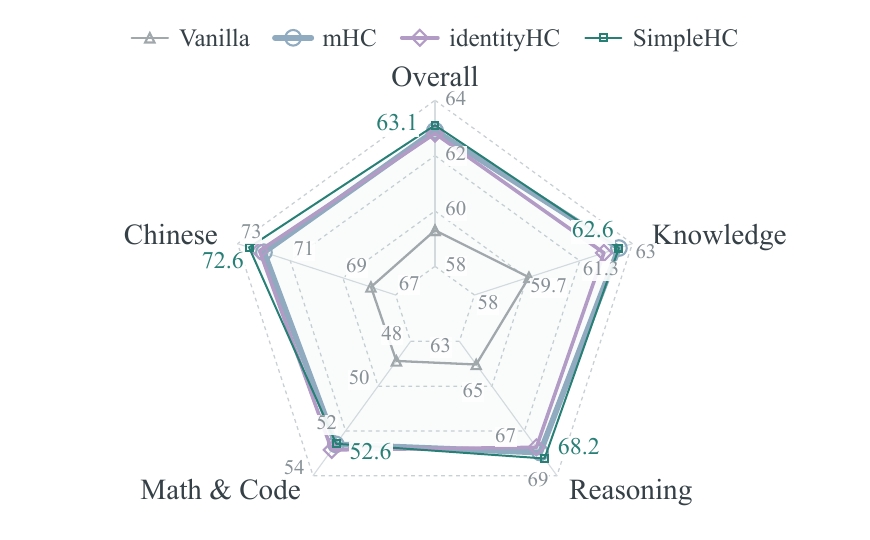}
  \par
  {\small (c) Downstream performance}
\end{minipage}

\caption{\textbf{SimpleHC reduces training overhead while retaining competitive performance.}
(a) Additional training overhead relative to baseline with
efficient implementations. (b) Storage retained for backward at
sublayer boundaries (Section~\ref{sec:exp-net}). (c) Average scores
on 30B MoE models (Table~\ref{tab:flagship_results}).}
\label{fig:simplehc-overview}

\end{figure}

Residual connections support optimization and information propagation
in modern language models
\citep{he2016deep,he2016identity,vaswani2017attention,grattafiori2024llama,yang2025qwen3,liu2024deepseekv3},
and recent work improves language modeling and depth utilization
by enriching this pathway
\citep{pagliardini2024denseformer,xiao2025muddformer,team2026attention_residual,qin2026multi}.
Hyper-Connections (HC)~\citep{zhu2025hyper} introduce multiple residual
streams with learned read, write, and mixing operations for flexible
information routing.
The manifold-constrained variant, mHC~\citep{xie2025mhc}, constrains
residual mixing to improve training stability at scale.
HC-based architectures, particularly mHC, have been adopted in
frontier language models~\citep{xu2026deepseek,GLM53flash}.
Despite the growing adoption, these architectures remain insufficiently
understood and costly to train: their gating and mixing operations
introduce architectural complexity and non-negligible execution
overhead, even under efficient implementations~\citep{xie2025mhc}.

A key source of this overhead is repeated access to the expanded residual
state: at every sublayer, HC-style mappings read and write $n$ streams
of width $C$ to generate gates, aggregate inputs, and propagate updates.
This memory traffic incurs costs not captured by FLOPs alone. In our end-to-end experiments with an efficient
implementation, mHC adds $7\%-27\%$ training time over a vanilla baseline
(Figure~\ref{fig:training-efficiency}(a)), despite adding less than $1\%$ FLOPs.
This motivates reconsidering the underlying read, write, and mixing
interactions. We ask:
\textbf{Can a simpler Hyper-Connections design reduce execution cost
while preserving model quality and training stability at scale?}

In this work, we systematically study the Hyper-Connections architecture through three
components: read gates, write gates, and residual mixing.
We find that unconstrained mixing improves quality, but Sinkhorn constraints
substantially reduce its benefit. Learned read/write gating without
mixing and unconstrained mixing with fixed gates each achieve quality
comparable to mHC, with further gains from combining them.
Our analysis also highlights the propagation risks of unconstrained
mixing, leading us to select learned read/write gating with identity
transitions as a minimal effective structure that retains competitive
quality while avoiding these risks.

Without residual mixing, the additive structure lets \emph{scalar corrections}
exactly recover preceding output contributions for given gates, without
materializing intermediate $nC$-dimensional states.
Gate generation, however, still accesses the updated state at every sublayer.
We reduce these accesses by computing layer-specific gates from a common
group-entry state, consolidating reads and deferring writes across sublayers.
We call this execution scheme \emph{shared memory access}.

Together, these findings lead to \textbf{SimpleHC}, which
\textbf{simplifies the architecture} through learned read/write gates
without mixing and \textbf{simplifies execution} through shared memory
access. We also develop specialized kernels to translate this simplicity into training efficiency.
As shown in Figure~\ref{fig:simplehc-overview}(a), SimpleHC incurs only
$2.4\%$ additional training cost at the largest tested scale
and halves memory traffic from $18C$ to $9C$
per sublayer relative to fused mHC.
It achieves downstream performance comparable
to mHC and identityHC on 15B and 30B MoE models
(Figure~\ref{fig:simplehc-overview}(c)). Analyses in Section~\ref{sec:discussion} further support the design of SimpleHC.
It also exhibits smoother gradient dynamics during
training and remains stable under stress-test settings.

\section{Preliminaries}

\label{sec:background}

Hyper-Connections (HC, dynamic variant)~\citep{zhu2025hyper} and Manifold-Constrained HC (mHC)~\citep{xie2025mhc}
share a multi-stream residual structure.  The two methods differ in how their coefficients are conditioned
and parameterized, and whether residual mixing is constrained.

Each Transformer block contains two \emph{sublayers}:
attention and FFN/MoE.
Let $X_l\in\mathbb{R}^{n\times C}$ contain $n$ width-$C$ residual streams
entering sublayer $l$. All vectors are column vectors;
$\operatorname{vec}(X_l)\in\mathbb{R}^{nC\times1}$ concatenates streams
in row order. We omit token and batch indices unless stated;
$I$ denotes the $n\times n$ identity matrix.
Both methods compute $u_l=X_l^\top h_l^{\mathrm{pre}}$,
$o_l=F_l(u_l)$, and
$X_{l+1}=H_l^{\mathrm{res}}X_l+h_l^{\mathrm{post}}o_l^\top$.
The read and write gates
$h_l^{\mathrm{pre}},h_l^{\mathrm{post}}\in\mathbb{R}^{n\times1}$
aggregate stream states and distribute sublayer outputs;
$H_l^{\mathrm{res}}\in\mathbb{R}^{n\times n}$ mixes streams.
Here, $u_l,o_l\in\mathbb{R}^{C\times1}$, and $F_l$ denotes the
sublayer module.

HC projects each normalized $C$-dimensional stream using
$W_l^{\mathrm{pre}},W_l^{\mathrm{post}}\in\mathbb{R}^{C\times1}$ and
$W_l^{\mathrm{res}}\in\mathbb{R}^{C\times n}$, shared across streams
within each sublayer. Scaled $\tanh$ projections are added to learned
static coefficients, yielding read/write gates and unconstrained
residual mixing.

In contrast, mHC uses the normalized full state
$z_l=\operatorname{RMSNorm}(\operatorname{vec}(X_l))\in\mathbb{R}^{nC\times1}$
to compute $\ell_l^s=\alpha_l^s(W_l^s)^\top z_l+b_l^s$
for $s\in\{\mathrm{pre},\mathrm{post},\mathrm{res}\}$.
Here,
$W_l^{\mathrm{pre}},W_l^{\mathrm{post}}\in\mathbb{R}^{nC\times n}$
and $W_l^{\mathrm{res}}\in\mathbb{R}^{nC\times n^2}$.
Each $\alpha_l^s$ is a learned scalar; logits and learned biases have
$n$ entries for read/write gates and $n^2$ for mixing.
Parameters are independent across operations and sublayers.
mHC applies sigmoid to read logits, twice sigmoid to write logits,
and Sinkhorn normalization to mixing logits reshaped into an
$n\times n$ matrix. Sinkhorn approximately enforces a doubly stochastic
$H_l^{\mathrm{res}}$ to stabilize propagation across depth. Unless otherwise stated, we use the full-state
gate parameterization of mHC.

\vspace{-5pt}
\section{SimpleHC}
\vspace{-2pt}

\label{sec:method}

\begin{figure}[!t]
\centering
\includegraphics[width=0.9\textwidth]{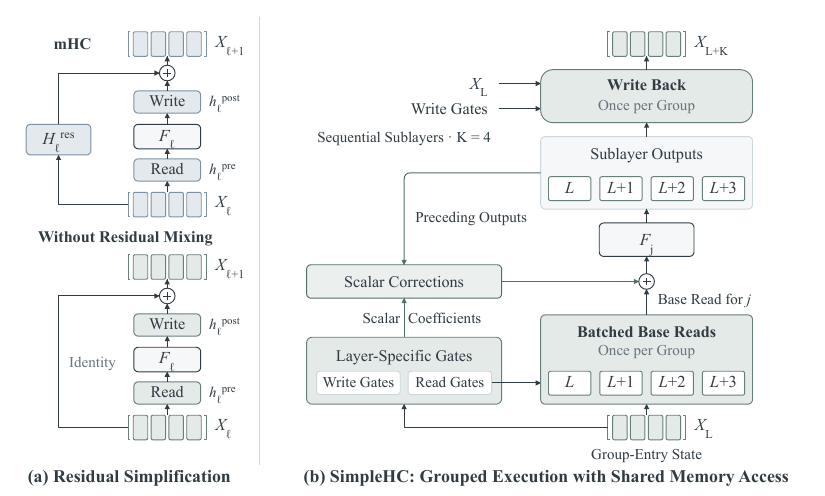}

\caption{\textbf{Overview of SimpleHC.}
(a) Identity transitions replace residual mixing, retaining learned
read/write gates.
(b) Layer-specific gates share a group-entry state across $K=4$
sublayers, enabling batched reads and one group-boundary write-back.
Scalar corrections preserve sequential output contributions
without materializing intermediate expanded states.}
\label{fig:simplehc-design}

\end{figure}

We propose SimpleHC, a Hyper-Connections architecture with a simplified
residual structure and execution scheme (Figure~\ref{fig:simplehc-design}).
It retains multiple residual streams and learned read/write gates
while removing residual mixing. Shared memory access reduces repeated
state access by generating layer-specific gates from group-entry
states, batching reads, and deferring writes.

\subsection{Simplifying the Residual Structure}

\label{sec:simple-structure}

Prior work observed that removing residual mixing from mHC by setting
$H_l^{\mathrm{res}}=I$ (identityHC) largely preserves model quality~\citep{guo2026ohc}.
identityHC has also been adopted in
Hy4 preview~\citep{tencent2026hy4preview}.
Building on these insights, our systematic study reveals an important distinction:
unconstrained residual mixing improves model quality, whereas
Sinkhorn-constrained mixing offers little benefit over identity transitions
(Table~\ref{tab:map}). The gains from unconstrained mixing, however,
come with the risk of cumulative signal amplification or attenuation
across depth
(Section~\ref{sec:analysis-minimal}).

We therefore retain learned read/write gates with identity transitions as a minimal
effective structure, preserving competitive quality while avoiding
residual matrix products, with
$X_{l+1}=X_l+h_l^{\mathrm{post}}o_l^\top$.
With $z_l=\operatorname{RMSNorm}(\operatorname{vec}(X_l))$ and
$\ell_l^s=\alpha_l^s(W_l^s)^\top z_l+b_l^s$
for $s\in\{\mathrm{pre},\mathrm{post}\}$, SimpleHC uses
$h_l^{\mathrm{pre}}=\sigma(\ell_l^{\mathrm{pre}})$ and
$h_l^{\mathrm{post}}=2\sigma(\ell_l^{\mathrm{post}})$. Without residual mixing, updates accumulate additively. Given the state
$X_L$ at sublayer $L$, the residual state at any later sublayer $j$ is
\begin{equation}
 X_j=X_L+\sum_{k=L}^{j-1}h_k^{\mathrm{post}}o_k^\top.
 \label{eq:accumulated-state}
\end{equation}
Each term adds a preceding sublayer's output to the streams according to its
write gate, and subsequent residual transitions leave this allocation
unchanged.

\subsection{Shared Memory Access}

\label{sec:share}

\begin{algorithm}[!t]
\caption{SimpleHC forward pass for one group}
\label{alg:shared-schedule}
\begin{algorithmic}[1]
\Require Entry state $X_L\in\mathbb{R}^{n\times C}$; group size $K$;
         gate functions $g_j^{\mathrm{pre}},g_j^{\mathrm{post}}$ and
         sublayers $F_j$ for $L\leq j<L+K$
\Ensure Group-boundary state $X_{L+K}$
\State $\bar H^{\mathrm{pre}}\gets[g_L^{\mathrm{pre}}(X_L),\ldots,
                  g_{L+K-1}^{\mathrm{pre}}(X_L)]$
\State $\bar H^{\mathrm{post}}\gets[g_L^{\mathrm{post}}(X_L),\ldots,
                  g_{L+K-1}^{\mathrm{post}}(X_L)]$
       \Comment{Independent gate columns, shared input}
\State $V\gets X_L^\top\bar H^{\mathrm{pre}}$ \Comment{Batched base reads; $V\in\mathbb{R}^{C\times K}$}
\State $D\gets\operatorname{tril}((\bar H^{\mathrm{pre}})^\top\bar H^{\mathrm{post}},-1)$
       \Comment{Strictly lower-triangular read/write coefficients}
\For{$a=0,\ldots,K-1$}
    \State $u_{L+a}\gets V_{:,a}+\sum_{b=0}^{a-1}D_{a,b}\,o_{L+b}$
           \Comment{Empty sum is zero}
    \State $o_{L+a}\gets F_{L+a}(u_{L+a})$
\EndFor
\State $O\gets[o_L,\ldots,o_{L+K-1}]$
\State $X_{L+K}\gets X_L+\bar H^{\mathrm{post}}O^\top$ \Comment{Materialize the group-boundary state}
\State \Return $X_{L+K}$
\end{algorithmic}
\end{algorithm}

Even without residual mixing, conventional execution accesses the
$nC$-dimensional state at every sublayer to generate gates, form inputs,
and write updates. We introduce shared memory access across sublayers
to reduce this traffic. With additive updates, each preceding output contributes to a later
input through the inner product of its write gate and the later read
gate. Substituting Equation~\ref{eq:accumulated-state} into the read
operation gives:
\begin{equation}
 u_j=X_j^\top h_j^{\mathrm{pre}}
 =X_L^\top h_j^{\mathrm{pre}}
 +\sum_{k=L}^{j-1}
 \underbrace{\bigl((h_j^{\mathrm{pre}})^\top
 h_k^{\mathrm{post}}\bigr)}_{\text{scalar coefficient}}o_k.
 \label{eq:rank-one-read}
\end{equation}
We call $X_L^\top h_j^{\mathrm{pre}}$ the \emph{base read} and the
remaining terms \emph{scalar corrections}, each weighting a preceding
$C$-dimensional output by a scalar coefficient.
For given gates, this reconstructs the input without writing intermediate
updates into the $nC$-dimensional residual state.

This reconstruction assumes that the gates are already available.
However, computing $h_j^s=g_j^s(X_j)$ requires materializing and accessing
the updated state at every sublayer, where $g_j^s$ is the gate mapping
defined above for $s\in\{\mathrm{pre},\mathrm{post}\}$.
SimpleHC relaxes this dependency by generating gates for $K$ consecutive
sublayers from a common group-entry state $X_L$:
\begin{equation}
 \bar h_j^{\mathrm{pre}}=g_j^{\mathrm{pre}}(X_L),\qquad
 \bar h_j^{\mathrm{post}}=g_j^{\mathrm{post}}(X_L),\qquad
 L\leq j<L+K.
 \label{eq:shared-input-gates}
\end{equation}
Within groups, sublayers retain independent gate parameters but
share the conditioning state $X_L$.
Relative to per-sublayer conditioning, the only approximation
is generating gates from $X_L$ rather than $X_j$.
Grouped gate generation computes layer-specific gates jointly
from $X_L$, reducing state reads from $K$ to $1$ and making group gates available upfront. With these gates, Equation~\ref{eq:rank-one-read} becomes
\begin{equation}
 u_j=X_L^\top\bar h_j^{\mathrm{pre}}
       +\sum_{k=L}^{j-1}d_{j,k}o_k,\qquad
 d_{j,k}=(\bar h_j^{\mathrm{pre}})^\top\bar h_k^{\mathrm{post}}.
 \label{eq:scalar-reconstruction}
\end{equation}
All $K$ base reads are batched into one scan of $X_L$, with
$d_{j,k}$ precomputed at group entry. Corrections incorporate
preceding $C$-dimensional outputs without intermediate
$nC$-dimensional state writes.

Within each group, sublayers execute sequentially, passing each
$o_j=F_j(u_j)$ to later inputs through scalar corrections.
A single fused kernel writes
back all updates at the group boundary:
\begin{equation}
 X_{L+K}=X_L+\sum_{k=L}^{L+K-1}\bar h_k^{\mathrm{post}}o_k^\top.
 \label{eq:group-boundary-write}
\end{equation}
This execution is algebraically equivalent to sequential execution
with the same group-entry gates
(Appendix~\ref{app:execution-proof}). SimpleHC uses $K=4$, where shared memory access
reduces full-state reads for gate generation and input aggregation
from 8 to 2 and writes from 4 to 1
relative to original execution.
Each full-state read/write accesses $nC$ elements per token.
Algorithm~\ref{alg:shared-schedule} summarizes the forward execution
of one group. Matrix columns use zero-based offsets within the group.

\section{Infrastructure Design}

\label{sec:infra}

\begin{table}[!t]
\centering
\caption{Memory access of residual-stream operations (reads and writes), per token per sublayer.}
\vspace{5pt}
\label{tab:infra-traffic}
\small
\setlength{\tabcolsep}{5pt}
\renewcommand{\arraystretch}{1.13}
\resizebox{0.9\textwidth}{!}{%
\begin{tabular}{lcccc}
\toprule
Operation & mHC (naive) & mHC (fused)
& identityHC (fused) & SimpleHC (fused) \\
\midrule
Projection and RMS statistics & $nC$ & $nC$ & $nC$ & $nC/K$ \\
Base reads & $nC+C$ & $nC+C$ & $nC+C$ & $nC/K+C$ \\
Residual write-back & $6nC+C$ & $2nC+C$ & $2nC+C$ & $2nC/K+C$ \\
\midrule
Total & $8nC+2C$ & $4nC+2C$ & $4nC+2C$ & $4nC/K+2C+Q_{\mathrm{corr}}(K)$ \\
$n=K=4$ & $34C$ & $18C$ & $18C$ & $9C$ \\
\bottomrule
\end{tabular}%
}

\end{table}

To realize the efficiency benefits of the simplified design,
we optimize the forward and backward execution of SimpleHC
with specialized kernels. At the largest tested width ($C=6144$),
these kernels reduce the additional end-to-end training time
over the vanilla residual baseline from $22.5\%$ with PyTorch
to $2.4\%$ (Figure~\ref{fig:training-efficiency}).
These kernels are implemented using DeepGEMM~\citep{deepgemm2025}
and Triton~\citep{tillet2019triton}.

\subsection{Forward Fusion with Shared State Access}
\label{sec:infra-gates}
\label{sec:infra-application}

\textbf{Grouped reads.}
Grouped gate generation and batched base reads implement the read side of
shared memory access. For
$x=\operatorname{vec}(X_L)\in\mathbb{R}^{nC\times1}$ and independent gate
projections concatenated into $W\in\mathbb{R}^{nC\times2Kn}$, we compute
\begin{equation}
 p=W^\top x,\qquad s=x^\top x,\qquad
 V=X_L^\top\bar H^{\mathrm{pre}}.
 \label{eq:infra-joint-projection}
\end{equation}
Here $p\in\mathbb{R}^{2Kn\times1}$, $s$ is scalar, and
$V\in\mathbb{R}^{C\times K}$. One kernel computes $p,s$ from the same
loaded state. Gate normalization uses affine-free RMSNorm, with
$r=(s/(nC)+\epsilon)^{-1/2}$ computed from the input state.
For each sublayer $j$ and gate type $t\in\{\mathrm{pre},\mathrm{post}\}$,
the corresponding projection block $p_j^t$ gives logits
$\ell_j^t=\alpha_j^t r p_j^t+b_j^t$.
A second kernel fuses this scaling and bias addition with gate activations,
avoiding a normalized-state buffer.
At $n=K=4$, one $4C\times32$ projection supplies four read gates and
four write gates. A separate batched-read kernel reuses each loaded state
tile for all $K$ distinct bases, storing them contiguously per sublayer.
Gate generation and base aggregation each reduce full-state reads from
$KnC$ to $nC$ elements per token per group.

\textbf{Fused corrections and deferred write-back.}
We implement Equations~\ref{eq:scalar-reconstruction}
and~\ref{eq:group-boundary-write} without stacked output copies or expanded
update buffers. Using precomputed coefficients $D$, each correction kernel
loads preceding $C$-dimensional outputs through separate pointers and fuses
weighting, summation, and base addition. At $K=4$, three correction kernels
supply the sequential sublayer inputs; one group-exit kernel then fuses
all four output allocations with residual addition. Wide-state update
traffic falls from $2KnC$ to $2nC$ elements per token per group.

\subsection{Backward Reuse and Efficient Reductions}
\label{sec:infra-backward}

We tailor backward kernels to two distinct workloads: wide activation
gradients benefit from cross-sublayer reuse, whereas small gate gradients
require efficient channel reductions.

For loss $\mathcal{L}$, let $G=\nabla_V\mathcal{L}\in\mathbb{R}^{C\times K}$,
$O=[o_L,\ldots,o_{L+K-1}]\in\mathbb{R}^{C\times K}$, and
$E=\nabla_{X_{L+K}}\mathcal{L}\in\mathbb{R}^{n\times C}$.
These equations give gradient contributions from the base reads
and final write-back, with $\mathrel{+}=$ denoting accumulation.

\textbf{Shared activation-gradient access.}
We compute activation-gradient contributions jointly across the group:
\begin{equation}
 \nabla_{X_L}\mathcal{L}\mathrel{+}=E+\bar H^{\mathrm{pre}}G^\top,
 \qquad
 \nabla_O\mathcal{L}\mathrel{+}=E^\top\bar H^{\mathrm{post}}.
 \label{eq:infra-write-backward}
\end{equation}
The read backward fuses all $K$ contributions with the identity gradient
$E$, avoiding $K$ expanded gradients and merging. Reusing one scan
of $E$ for all $K$ output gradients reduces its logical reads from $KnC$
to $nC$ elements per token per group ($4\times$ at $K=4$).
We fuse projection input gradients with RMS differentiation, avoiding
a wide intermediate and a separate normalization-backward pass;
weight gradients use a separate GEMM.

\textbf{Token-packed gate gradients.}
Gate gradients instead require reductions over the channel dimension:
\begin{equation}
 \nabla_{\bar H^{\mathrm{pre}}}\mathcal{L}\mathrel{+}=X_LG,
 \qquad
 \nabla_{\bar H^{\mathrm{post}}}\mathcal{L}\mathrel{+}=EO.
 \label{eq:infra-gate-backward}
\end{equation}
At $n=K=4$, each product reduces $C$ channels to just 16 coefficients
per token. We pack four tokens into a $16\times16$ Tensor Core product,
accumulate over channels on-chip, and store only the four same-token
$4\times4$ blocks, eliminating the partial-gradient scratch buffer and
extra reduction launch of the channel-tiled fallback for each product.

\subsection{Memory-Access and Storage Accounting}
\label{sec:infra-accounting}

The accounting in Table~\ref{tab:infra-traffic} covers logical forward
activation reads and writes per token per sublayer, excluding parameters,
small coefficient buffers, and attention/FFN internals. Naive mHC
materializes mixed states and expanded updates separately; fused mHC
combines residual mixing, output allocation, and residual addition.

At group offset $a>0$, each correction reads one base and $a$ preceding
outputs, then writes one input, costing $(a+2)C$ elements. Offset $a=0$
uses its base directly. Averaging over $K$ sublayers gives
\begin{equation}
 Q_{\mathrm{corr}}(K)=\frac{C}{K}\sum_{a=1}^{K-1}(a+2)
 =\frac{C(K-1)(K+4)}{2K},\qquad Q_{\mathrm{corr}}(4)=3C.
 \label{eq:infra-correction-traffic}
\end{equation}
Grouped gate generation and base reads each access $nC$ elements per
group; final write-back reads and writes the wide state once, adding
$2nC$. Together these contribute $4nC/K$ per sublayer. Base writes and
output reads for final write-back add $2C$, yielding
\begin{equation}
 Q_{\mathrm{SimpleHC}}(K)=\frac{4nC}{K}+2C+Q_{\mathrm{corr}}(K).
 \label{eq:infra-total-traffic}
\end{equation}
At $n=4$, this gives $11.5C$, $9C$, $9.25C$, and $12.375C$ for
$K=2,4,8,16$, respectively. Thus, $K=4$ has the lowest logical traffic
among these group sizes, rather than over all positive integers.
Correction costs grow with $K$, offsetting
the reduced wide-state traffic at larger group sizes.
Our choice of $K=4$ halves the $18C$ traffic of fused mHC
and identityHC. Removing mixing alone does not reduce these accesses:
fused mHC already combines mixing and output allocation in one write-back.
The reduction comes from sharing state access across sublayers. Without recomputation, saving one group-entry residual state
instead of $K$ sublayer-entry states reduces the storage of
these residual states by $1-1/K$ ($75\%$ at $K=4$).

\section{Experiments}

\label{sec:exp}

\begin{table}[!t]
\centering
\caption{\textbf{Architectural design choices in Hyper-Connections.}
Using 15B-A1B MoE models. $C$/$nC$ denote stream-local/joint-state gate
inputs; Fixed uses uniform weights of $1/n=1/4$ for read gates and $1$ for write gates.}
\vspace{5pt}
\label{tab:map}
\begingroup
\footnotesize
\setlength{\tabcolsep}{2.5pt}
\renewcommand{\arraystretch}{1.08}
\resizebox{\textwidth}{!}{%
\begin{tabular}{@{\hspace{4pt}}lccc|ccccccc@{\hspace{4pt}}}
\toprule
\textbf{Method} & Read & Write & Mixing
& Loss $\downarrow$ & MMLU & HellaSwag
& MATH & MBPP & C3 & Avg. \\
\midrule
(1) Vanilla Residual & --- & --- & --- & 1.678 & 56.1 & 65.8 & 26.2 & 37.8 & 75.9 & 52.4 \\
\midrule
\multicolumn{11}{c}{\textit{Gate inputs and residual mixing}} \\
\cmidrule{1-11}
(2) HC & $C$ & $C$ & Unconstrained & 1.660 & \textbf{60.9} & 66.4 & \textbf{30.4} & 39.8 & 81.6 & 55.8 \\ %
(3) HC w/o mixing & $C$ & $C$ & Identity & 1.664 & 59.6 & 65.5 & 30.1 & 37.6 & 82.9 & 55.1 \\ %
(4) mHC w/o Sinkhorn & $nC$ & $nC$ & Unconstrained & \textbf{1.657} & 60.0 & 66.4 & 27.9 & \textbf{42.6} & \textbf{84.2} & \textbf{56.2} \\ %
(5) mHC & $nC$ & $nC$ & Sinkhorn & 1.659 & 58.8 & \textbf{67.3} & 26.9 & 39.8 & 82.2 & 55.0 \\ %
(6) identityHC & $nC$ & $nC$ & Identity & 1.660 & 59.6 & 66.4 & 29.6 & 39.6 & 81.9 & 55.4 \\ %
\midrule
\multicolumn{11}{c}{\textit{Read- and write-input restrictions}} \\
\cmidrule{1-11}
(7) Restricted read & $C$ & $nC$ & Identity & \textbf{1.661} & \textbf{59.9} & \textbf{66.9} & 28.5 & 39.0 & \textbf{80.1} & \textbf{54.9} \\ %
(8) Restricted write & $nC$ & $C$ & Identity & 1.663 & 59.1 & 66.4 & \textbf{29.0} & \textbf{39.2} & 79.9 & 54.7 \\ %
\midrule
\multicolumn{11}{c}{\textit{Fixed-gate controls}} \\
\cmidrule{1-11}
(9) Fixed read & Fixed & $nC$ & Identity & 1.676 & 57.5 & 65.5 & 27.9 & 38.4 & 78.5 & 53.6 \\ %
(10) Fixed write & $nC$ & Fixed & Identity & 1.678 & 56.2 & 64.3 & 27.5 & 38.0 & 78.1 & 52.8 \\ %
(11) Fixed read and write & Fixed & Fixed & Sinkhorn & 1.678 & 56.5 & 64.7 & 26.7 & 37.4 & 77.0 & 52.5 \\ %
(12) Fixed read and write & Fixed & Fixed & Unconstrained & \textbf{1.660} & \textbf{61.9} & \textbf{67.1} & \textbf{29.1} & \textbf{40.2} & \textbf{81.8} & \textbf{56.0} \\ %
\bottomrule
\end{tabular}%
}
\endgroup

\end{table}

\subsection{Setup}

\label{sec:setup}
\textbf{Training setup.}
We conduct experiments on a MoE model with 15B total and
1B activated parameters (15B-A1B), and evaluate
scalability on a 30B-A2B model.
Following dots3-note Preview~\citep{dots2026dots3note}, we use
MoE backbones~\citep{dai2024deepseekmoe} with MLA~\citep{liu2024deepseekV2_MLA},
alternating one full-attention layer with three sliding-window layers.
We train with a sequence length of 8192 using Muon \citep{liu2025muon}
on a high-quality mixture of English and Chinese text, code, and mathematics.
Unless specified, we use $n=4$ streams and groups of
$K=4$ sublayers.
Appendix~\ref{app:training-config} details configurations.

\textbf{Evaluation.}
We evaluate knowledge, reasoning, math, code, and Chinese on
MMLU \citep{MMLU}, HellaSwag \citep{zellers2019hellaswag},
MATH \citep{MATH}, MBPP \citep{MBPP}, and C3 \citep{C3}
for the 15B-A1B studies (Tables~\ref{tab:map} and~\ref{tab:share}).
For 30B-A2B (Table~\ref{tab:flagship_results}), we add
MMLU-Pro \citep{wang2024mmlupro}, MMLU-Redux \citep{MMLURedux},
TriviaQA \citep{joshi2017triviaqa}, DROP \citep{dua2019drop},
BBH \citep{BBH}, GSM8K \citep{GSM8k}, HumanEval \citep{HumanEval},
CMMLU \citep{li2024cmmlu}, and CEval \citep{ceval},
yielding fourteen benchmarks.
Appendix~\ref{app:eval-protocols} details few-shot settings and loss reporting
for both the 15B-A1B and 30B-A2B models.

\subsection{Understanding the Hyper-Connections Architecture}

\label{sec:map}

We examine the roles and interactions of read/write gates
and residual mixing in the HC architecture (Table~\ref{tab:map}).
HC and mHC share these components but differ in gate conditioning,
parameterization, and mixing constraints.
Except for HC variants, gates follow the parameterization in
Section~\ref{sec:simple-structure}.

\textbf{Configurations.}
HC w/o mixing retains the HC parameterization but sets
$H^{\mathrm{res}}=I$; identityHC makes the same replacement in mHC.
mHC w/o Sinkhorn and the unconstrained fixed-gate control use
$H^{\mathrm{res}}=I+\operatorname{reshape}(\ell^{\mathrm{res}})$,
without exponentiation or Sinkhorn normalization.
Restricted read/write replaces only the indicated gate's input with
the corresponding stream, using per-stream RMS normalization and a
$C\times1$ projection shared across streams; the other gate retains
joint-state conditioning.
Fixed read/write sets the indicated gate to $\mathbf{1}/n$ or
$\mathbf{1}$, respectively, while retaining the other components
specified in the table.

We observe:
(i) \textbf{Unconstrained mixing is effective, but Sinkhorn diminishes its gains.}
Retaining unconstrained mixing improves the average score in HC (rows 3 vs.\ 2). Under the mHC parameterization, unconstrained
mixing achieves 56.2, versus 55.0 with Sinkhorn constraints and 55.4
with identity transitions (rows 4--6). Sinkhorn limits the benefit
of mixing, leaving performance close to identityHC.
(ii) \textbf{Three components form two effective groups: read/write gating and unconstrained mixing.}
Without mixing, jointly learned read/write gates achieve an average
score of 55.4; fixing either gate lowers it significantly, supporting their treatment as one group.
Unconstrained mixing forms the other: with both gates fixed, it
achieves 56.0 (row 12). Either group alone is comparable to mHC;
combining them improves the score (row 4).
(iii) \textbf{Full-state conditioning improves model quality.}
\label{sec:asymmetry}
With full-state inputs, loss is 1.660.
Restricting read/write gates to stream-local inputs raises loss by
0.001/0.003 (rows 6--8), suggesting
slightly greater sensitivity in write gates.
Section~\ref{sec:analysis-minimal} analyzes these findings and
their implications for SimpleHC.

\subsection{Shared Memory Access with Layer-Specific Gating}

\label{sec:exp-share}

Table~\ref{tab:share} ablates group size, gate reuse, and scalar
corrections, with identityHC and mHC as references.

\textbf{Configurations.}
Share $K$ generates gates for $K$ consecutive sublayers from one
group-entry state; $K=4$ spans two Transformer blocks.
Independent gates retain separate parameters for each sublayer,
whereas reused gates share one read/write pair across the group.
The w/o corrections variant uses only base reads for sublayer inputs
without accounting for preceding sublayer output contributions,
but retains all sublayer outputs in the group-boundary write-back.
All sharing variants use identity residual transitions.

The results highlight three findings:
(i) \textbf{Moderate state sharing preserves model quality.}
$K=2,4$ match the loss of identityHC, while $K=8$ slightly increases
it (rows 3--6). At $K=4$, the average score of 55.0 is comparable to mHC and identityHC.
We choose $K=4$ for its competitive quality and lowest memory traffic (Section~\ref{sec:infra-accounting}).
(ii) \textbf{Layer-specific gates improve quality under shared inputs.}
At $K=4$, reusing one read/write gate pair lowers the average score
from 55.0 to 53.5 (rows 5 vs.\ 7). This supports retaining layer-specific gate parameters.
(iii) \textbf{Scalar corrections are important for model performance.}
Removing corrections raises loss from 1.660 to 1.705 despite retaining
all outputs in the final write-back (rows 5 vs.\ 8).
Thus, later sublayers must incorporate preceding outputs even when
gate inputs are shared.
Section~\ref{sec:analysis-sharing} analyzes these findings.

\begin{table}[!t]
\centering
\caption{\textbf{Design of shared memory access.}
$K$: sublayers per group.
Independent/Reused: layer-specific/shared read/write gates.
Row 8 omits corrections but retains group-boundary write-back.}
\vspace{5pt}
\label{tab:share}
\begingroup
\footnotesize
\setlength{\tabcolsep}{2.5pt}
\renewcommand{\arraystretch}{1.08}
\resizebox{\textwidth}{!}{%
\begin{tabular}{@{\hspace{4pt}}lcccc|ccccccc@{\hspace{4pt}}}
\toprule
\textbf{Method} & Memory access & Group size $K$ & Mixing & Gates
& Loss $\downarrow$ & MMLU & HellaSwag
& MATH & MBPP & C3 & Avg. \\
\midrule
(1) Vanilla Residual & --- & --- & --- & --- & 1.678 & 56.1 & 65.8 & 26.2 & 37.8 & 75.9 & 52.4 \\
\midrule
\multicolumn{12}{c}{\textit{Reference configurations}} \\
\cmidrule{1-12}
(2) mHC & Per sublayer & 1 & Sinkhorn & Independent & \textbf{1.659} & 58.8 & \textbf{67.3} & 26.9 & \textbf{39.8} & \textbf{82.2} & 55.0 \\
(3) identityHC & Per sublayer & 1 & Identity & Independent & 1.660 & \textbf{59.6} & 66.4 & \textbf{29.6} & 39.6 & 81.9 & \textbf{55.4} \\
\midrule
\multicolumn{12}{c}{\textit{Sharing granularity}} \\
\cmidrule{1-12}
(4) Share 2 & Shared & 2 & Identity & Independent & \textbf{1.660} & \textbf{59.9} & \textbf{67.0} & 28.1 & 39.0 & 80.1 & 54.8 \\ %
(5) SimpleHC (Share 4) & Shared & 4 & Identity & Independent & \textbf{1.660} & 58.5 & \textbf{67.0} & \textbf{28.7} & \textbf{40.8} & \textbf{80.2} & \textbf{55.0} \\
(6) Share 8 & Shared & 8 & Identity & Independent & 1.662 & 59.1 & 66.2 & 27.9 & 39.4 & 79.1 & 54.3 \\ %
\midrule
\multicolumn{12}{c}{\textit{Layer-specific gating}} \\
\cmidrule{1-12}
(7) Share 4, reused gates & Shared & 4 & Identity & Reused & 1.667 & 58.1 & 67.1 & 28.4 & 35.4 & 78.4 & 53.5 \\ %
\midrule
\multicolumn{12}{c}{\textit{Within-group update corrections}} \\
\cmidrule{1-12}
(8) Share 4 w/o corrections & Shared & 4 & Identity & Independent & 1.705 & 52.8 & 64.4 & 21.7 & 32.0 & 68.8 & 47.9 \\ %
\bottomrule
\end{tabular}%
}
\endgroup

\end{table}

\subsection{Evaluating SimpleHC at Larger Scale}

\label{sec:exp-scale}

On the larger 30B-A2B model, SimpleHC matches mHC performance
while simplifying both the residual structure and execution.
As shown in Table~\ref{tab:flagship_results}, it achieves the same loss of
1.561 as mHC and identityHC and an average benchmark score of 63.1 versus
62.9 for mHC and 62.8 for identityHC. It achieves higher scores on MMLU,
MATH, and C3, although it trails mHC on GSM8K and both methods on HumanEval. These results support the scalability of SimpleHC.

\begin{table}[!t]
\centering
\small
\caption{\textbf{Downstream evaluation on 30B MoE models.}
We compare Vanilla, mHC, identityHC, and SimpleHC across benchmarks covering Knowledge,
Reasoning, Math \& Code, and Chinese.}
\label{tab:flagship_results}
\setlength{\tabcolsep}{5pt}
\renewcommand{\arraystretch}{1.08}
\vspace{5pt}
\resizebox{0.72\textwidth}{!}{%
\begin{tabular}{@{\hspace{4pt}}llcccc@{\hspace{4pt}}}
\toprule
 & Benchmark & Vanilla & mHC & identityHC & SimpleHC \\
\midrule
 & Loss $\downarrow$ & 1.579 & \textbf{1.561} & \textbf{1.561} & \textbf{1.561} \\
\midrule
\multirow{4}{*}{\textit{Knowledge}} & MMLU & 65.8 & 67.7 & 67.9 & \textbf{68.6} \\
 & MMLU-Pro & 45.6 & 48.6 & 48.0 & \textbf{48.9} \\
 & MMLU-Redux & 62.4 & 64.3 & 64.1 & \textbf{64.7} \\
 & TriviaQA & 65.1 & \textbf{69.7} & 68.4 & 68.1 \\
\midrule
\multirow{3}{*}{\textit{Reasoning}} & HellaSwag & 71.3 & 72.4 & 72.8 & \textbf{73.0} \\
 & DROP & 61.4 & 67.2 & 67.7 & \textbf{68.1} \\
 & BBH & 59.4 & \textbf{64.3} & 62.7 & 63.6 \\
\midrule
\multirow{4}{*}{\textit{Math \& Code}} & GSM8K & 71.3 & \textbf{75.0} & 74.5 & 74.5 \\
 & MATH & 38.4 & 38.4 & 38.9 & \textbf{41.5} \\
 & HumanEval & 39.6 & 47.6 & \textbf{48.2} & 45.1 \\
 & MBPP & 46.2 & 49.4 & \textbf{49.8} & 49.2 \\
\midrule
\multirow{3}{*}{\textit{Chinese}} & CMMLU & 55.1 & \textbf{60.7} & 58.7 & 59.0 \\
 & CEval & 60.6 & 65.7 & \textbf{68.2} & 67.3 \\
 & C3 & 88.1 & 89.6 & 89.6 & \textbf{91.4} \\
\midrule
 & \textbf{Average} & 59.3 & 62.9 & 62.8 & \textbf{63.1} \\
\bottomrule
\end{tabular}%
}

\end{table}

\subsection{Translating Simplicity into Training Efficiency}

\label{sec:exp-net}

We compare end-to-end training overhead against the baseline
across scales from $C=1024$ to $C=6144$ (matching frontier models~\citep{xu2026deepseek,tencent2026hy4preview}).
Details are provided in Appendix~\ref{app:efficiency-settings}.

\begin{figure}[!t]
    \centering
    \begin{minipage}[t]{0.46\linewidth}
        \centering
        \includegraphics[width=\linewidth]{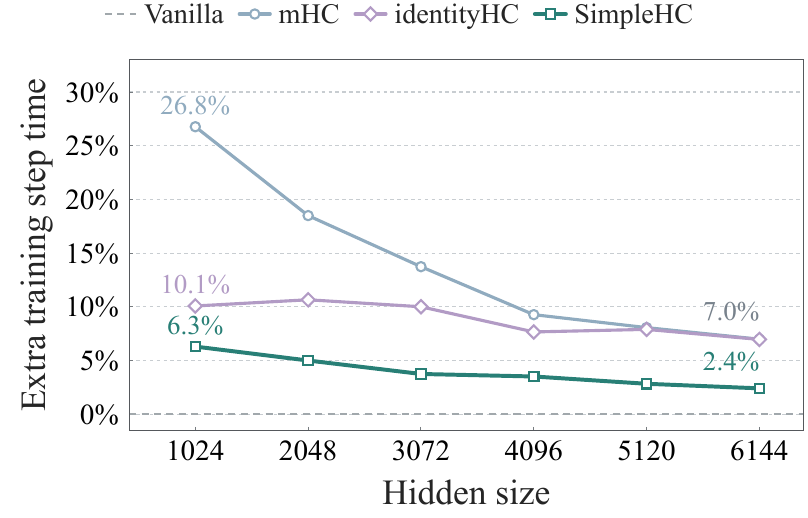}
        \par
        {\small (a) Efficient implementations}
    \end{minipage}\hfill
    \begin{minipage}[t]{0.46\linewidth}
        \centering
        \includegraphics[width=\linewidth]{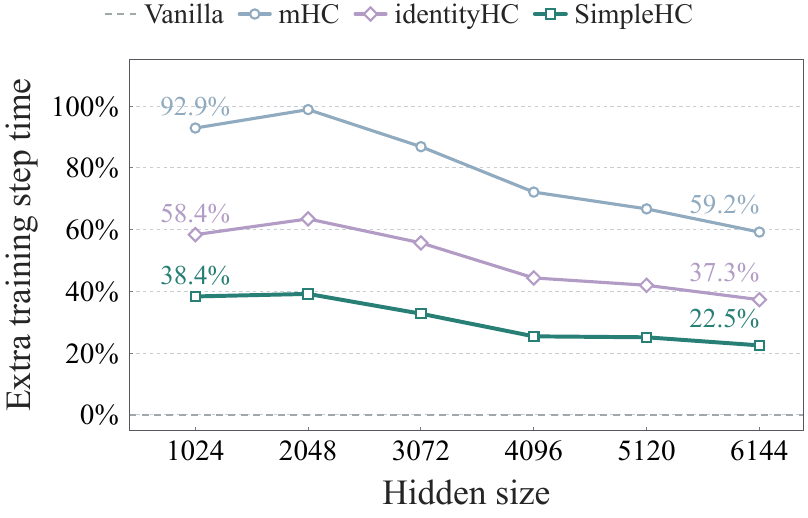}
        \par
        {\small (b) PyTorch implementations}
    \end{minipage}
    \caption{\textbf{End-to-end training overhead relative to Vanilla Residual.}
    SimpleHC achieves the lowest overhead across tested model widths
    with both (a) efficient and (b) PyTorch implementations.}
    \label{fig:training-efficiency}
\end{figure}

With the efficient implementations,
SimpleHC achieves the lowest overhead across all tested scales
(Figure~\ref{fig:training-efficiency}(a)).
At the smallest scale, overhead is 6.3\%, versus 26.8\% for mHC
and 10.1\% for identityHC.
Despite matching forward activation traffic, mHC incurs
additional mixing and Sinkhorn costs, explaining its larger gap
from identityHC at smaller scales.
Relative overhead generally decreases with width: HC operations
scale linearly at fixed $n$, whereas expert projections scale
quadratically when expert width grows proportionally.
At the largest tested scale, SimpleHC cuts overhead from approximately
7.0\% for both baselines to 2.4\%, eliminating about two-thirds.

\begin{figure}[!t]
    \centering
    \includegraphics[width=0.60\linewidth]{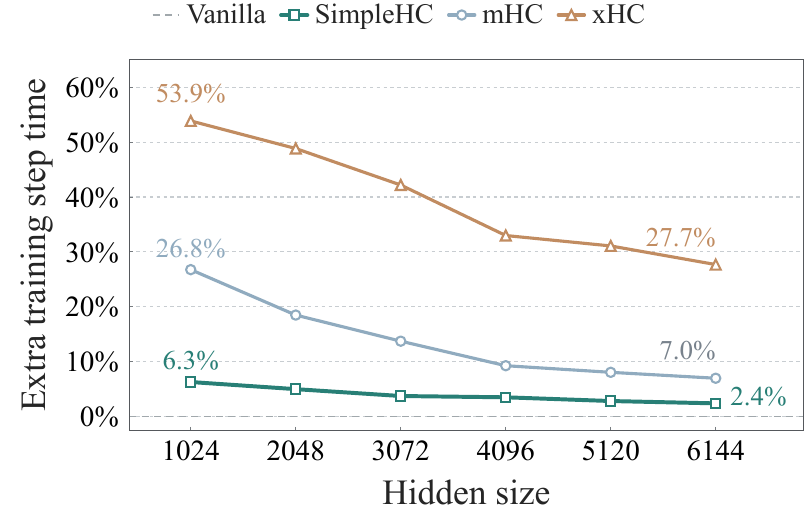}
    \caption{\textbf{Training overhead compared with mHC and xHC.}
    All methods use efficient implementations, with overhead measured
    relative to the vanilla residual baseline. xHC uses 16 residual
    streams with 4 active streams; mHC and SimpleHC use 4
    residual streams.}
    \label{fig:efficiency-xhc}
\end{figure}

We additionally compare against xHC, a promising approach to scaling
residual-stream capacity~\citep{zhang2026xhc}. At $C=6144$, its training
overhead is 27.7\%, compared with approximately 7.0\% for mHC and 2.4\%
for SimpleHC (Figure~\ref{fig:efficiency-xhc}). These results position SimpleHC as one of the most efficient HC variants
evaluated for scaling foundation models.

Specialized kernels require substantial engineering effort, limiting
wider adoption. Thus we also evaluate PyTorch versions
(Figure~\ref{fig:training-efficiency}(b)), where SimpleHC has the lowest
overhead across all tested scales.

\textbf{Saved activation memory.}
By reusing group-entry states, SimpleHC reduces estimated storage
retained for backward at sublayer boundaries and within HC connections
by approximately 31\% relative to mHC
(Figure~\ref{fig:simplehc-overview}(b)). Appendix~\ref{app:saved-activation-accounting}
details the accounting.

\let\savedFloatBarrier\FloatBarrier
\let\FloatBarrier\relax
\section{Analysis}
\let\FloatBarrier\savedFloatBarrier

\label{sec:mechanism-analysis}
\label{sec:discussion}

We analyze the design and training stability of SimpleHC.
Three conclusions emerge:
(i) \textbf{learned read/write gating provides a minimal effective
residual structure};
(ii) \textbf{gating decisions vary across sublayers yet remain predictable
from a shared group-entry state};
(iii) \textbf{SimpleHC remains stable under stress and exhibits smoother
gradient dynamics during standard training}.

\subsection{A Minimal Structure for Hyper-Connections}

\label{sec:analysis-minimal}

To combine architectural simplicity, competitive performance, and robust
scaling, we select learned read/write gating without mixing as a minimal
effective residual structure for Hyper-Connections. This avoids the
propagation risks of cumulative residual mixing, as analyzed below.

Learned read/write gating and unconstrained mixing form two effective
groups: each achieves performance comparable to mHC, with further gains
when combined (Table~\ref{tab:map}).
Appendix~\ref{app:interaction-groups} compares their weighting mechanisms.
However, unconstrained mixing repeatedly transforms earlier contributions
through residual matrix products, risking cumulative amplification
and training instability~\citep{xie2025mhc}.
We observe unconstrained mixing with fixed gates produces larger peak
residual-path coefficients than identityHC, consistent with these
amplification risks (Figure~\ref{fig:residual-transfer},
Appendix~\ref{app:transfer-diagnostics}).

Sinkhorn constraints mitigate these risks~\citep{xie2025mhc}, but
constrained mixing adds little benefit over learned read/write gating
alone. We therefore retain learned gates without mixing and use
full-state inputs for better loss and benchmark scores
(Table~\ref{tab:map}). Full-state and stream-local conditioning
incur the same residual-state input traffic, as both read all streams.

\subsection{Why Shared Memory Access Is Effective}

\label{sec:analysis-sharing}

\begin{figure}[!t]
\centering
\begin{minipage}[t]{0.48\textwidth}
\centering
\includegraphics[width=\linewidth]{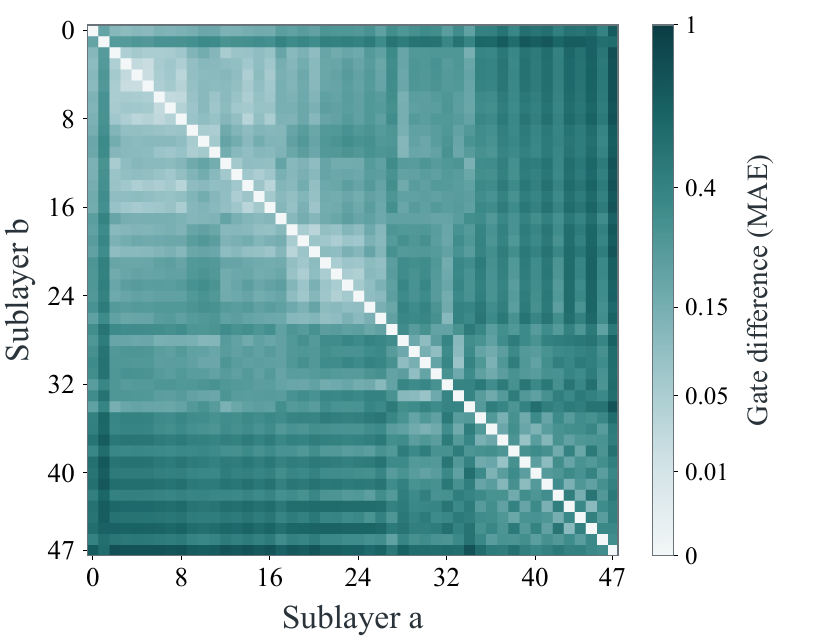}
\par
{\small (a) Write-gate variation across sublayers}
\end{minipage}\hfill
\begin{minipage}[t]{0.475\textwidth}
\centering
\includegraphics[width=\linewidth]{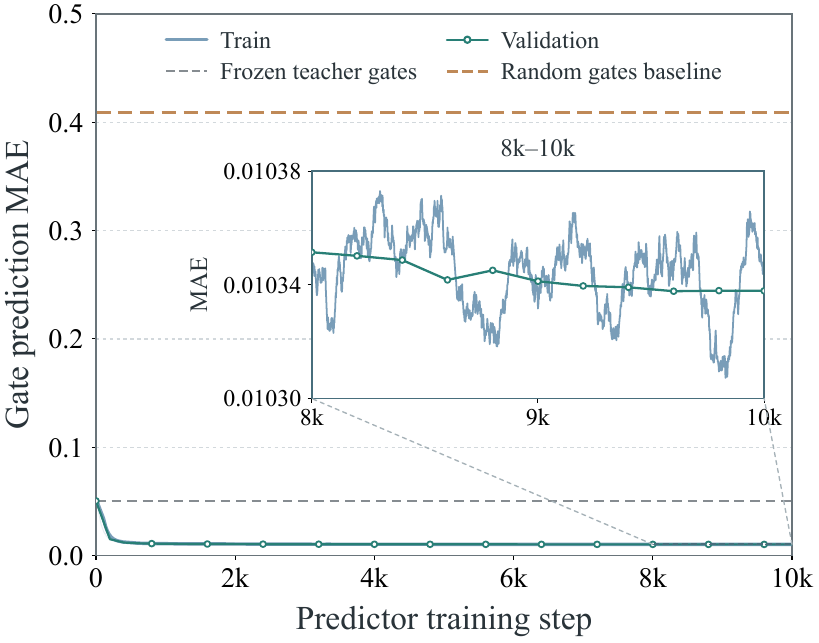}
\par
{\small (b) Prediction from the group-entry state}
\end{minipage}

\caption{\textbf{Gates vary across sublayers yet are predictable from shared inputs.}
(a) Pairwise gate MAE across identityHC sublayers, averaged over streams
and tokens.
(b) Train and val. MAE for write-gate prediction at offsets 1--3
from the group-entry state.
See Appendix~\ref{app:gate-prediction} for details.}
\label{fig:gate-sharing-analysis}

\end{figure}

We examine why shared memory access benefits from layer-specific gates
and remains effective with shared inputs through gate variation
and predictability analyses.

\textbf{Gate variation across sublayers.}
Read and write gates exhibit distinct allocation patterns across
sublayers of the 15B identityHC model
(Figures~\ref{fig:read-gate-analysis}(a)
and~\ref{fig:gate-sharing-analysis}(a)).
These differences motivate retaining layer-specific gates under shared
inputs, consistent with their advantage over gate reuse
(Table~\ref{tab:share}).

\textbf{Predictability from the group-entry state.}
We examine whether subsequent gates are predictable from the group-entry
state to assess the feasibility of sharing gate inputs.
We train layer-specific predictors to recover subsequent read/write gates
of a frozen 15B identityHC model from the group-entry state.
The predictors use the SimpleHC parameterization and are initialized
from the corresponding teacher gate parameters.
Only predictors are trained.
The ``Frozen teacher gates'' baseline uses these mappings without training.
In Figure~\ref{fig:gate-sharing-analysis}(b), it predicts write gates more
accurately than uniform random gates (MAE 0.0504 vs.\ 0.4086), indicating
predictive ability from earlier inputs even without adaptation.
Training further reduces MAE to 0.0103, suggesting that layer-specific
mappings closely approximate subsequent gates from shared inputs and
supporting the shared memory access design.
Read gates show the same trend (Figure~\ref{fig:read-gate-analysis}(b)). Appendix~\ref{app:gate-prediction}
details the protocol.

Sharing gate inputs does not remove the need for sequential output
contributions: removing scalar corrections raises loss from 1.660
to 1.705 despite retaining the final write-back.

\subsection{Does Simplification Compromise Training Stability?}

\label{sec:disc-stability}

\begin{figure}[!t]
\centering
\begin{minipage}[t]{0.49\textwidth}
\centering
\includegraphics[width=\linewidth]{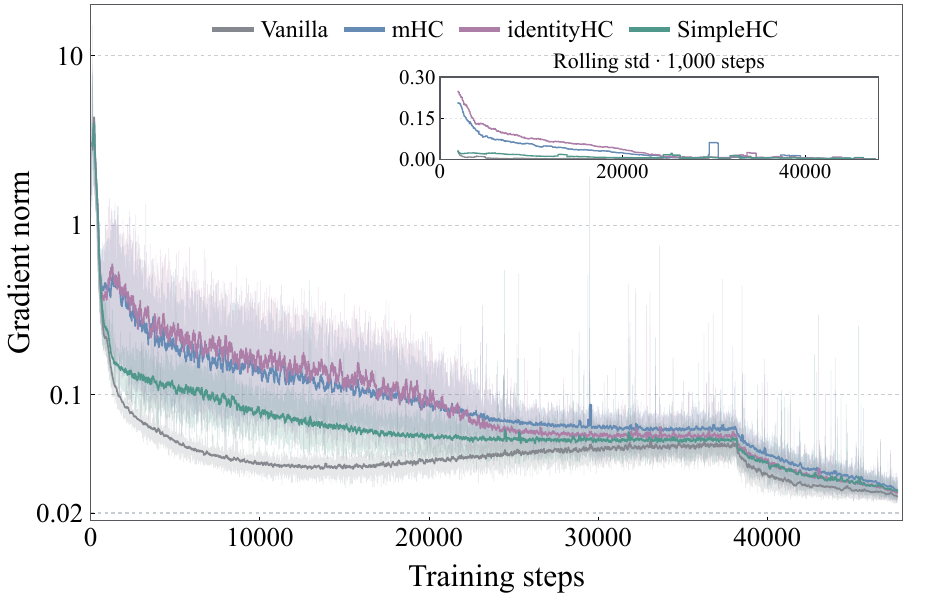}
\par
{\small (a) Standard 15B training}
\end{minipage}\hfill
\begin{minipage}[t]{0.49\textwidth}
\centering
\includegraphics[width=\linewidth]{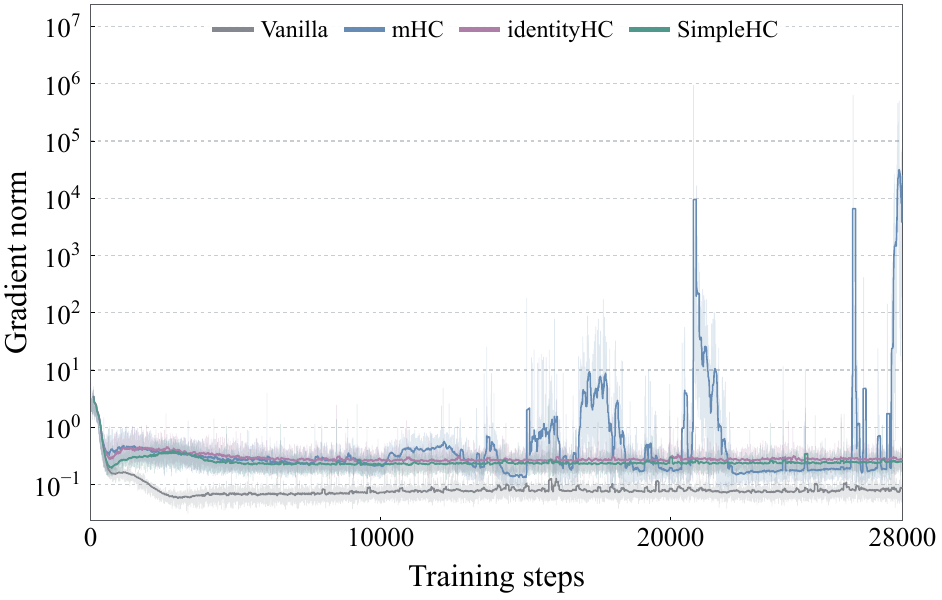}
\par
{\small (b) Stress test}
\end{minipage}

\caption{\textbf{Training stability of SimpleHC.}
(a) Inset: 1,000-step rolling gradient-norm std.
(b) Under stress, mHC exhibits severe gradient spikes, while SimpleHC
and identityHC remain stable.}
\label{fig:training-stability}

\end{figure}

Training stability is essential for reliable scaling, as instabilities
absent in smaller models can emerge at larger scales~\citep{wortsman2024small}.
We examine whether the simplifications in SimpleHC compromise
training stability.
During 15B training, SimpleHC shows smaller gradient-norm fluctuations
and lower rolling standard deviation than mHC and identityHC early
after warmup (Figure~\ref{fig:training-stability}(a)); all three reach
comparable ranges later.
Shared-input gating may reduce sensitivity to within-group state
fluctuations by conditioning decisions on an earlier state, loosely
analogous to stabilization through historical averaging.
This could contribute to smoother training of SimpleHC.

To emulate large-scale training stability challenges, we stress-test
230B-A6B MoE models at $10\times$ the reference learning rate.
In Figure~\ref{fig:training-stability}(b), mHC exhibits gradient spikes
reaching approximately $1.9\times10^6$, while SimpleHC and identityHC
maintain bounded gradient norms.
Appendix~\ref{app:stability-details} details the settings and analyzes
potential causes of mHC instability.

\section{Related Work}

\label{sec:related}

\textbf{Hyper-Connections and Variants.}
Hyper-Connections expand the residual pathway into multiple streams with
learned read, write, and mixing operations \citep{zhu2025hyper}.
Subsequent work constrains mixing to doubly stochastic matrices
\citep{xie2025mhc}, reduces constraint-enforcement costs
\citep{yang2026mhclite,zhou2026kromhc,dandachi2026gomHC},
or explores other constraints for stable propagation
and stream diversity \citep{liu2026sHC,guo2026ohc}.
Other designs modify read/write structures \citep{qiu2026qwen3.8,MGR}
or use sparse updates to support more streams \citep{zhang2026xhc}.

\textbf{Understanding and Simplifying Hyper-Connections.}
Recent analyses reveal uneven stream utilization, redundancy, and
concentrated read/write routing in HC models~\citep{peng2026ablateandrescue,stream_collapse}.
Prior work also reports that removing residual mixing can largely preserve
model quality~\citep{guo2026ohc}, with related evidence from
finetuning~\citep{mhcpeft}. These observations motivate examining
which interactions account for the benefits.
Our study identifies learned read/write gating and unconstrained mixing
as two independently effective groups, while showing that
Sinkhorn constraints diminish the gains from mixing.
These findings explain why learned gating alone remains competitive with mHC~\citep{xie2025mhc}
and motivate an additive structure whose execution we further simplify
through shared memory access.

\textbf{Efficient Execution of Residual Aggregation.}
Richer residual pathways introduce storage and data-movement costs beyond
their arithmetic overhead. Prior work reduces these costs through block-wise
aggregation and staged computation \citep{team2026attention_residual},
kernel fusion and recomputation \citep{xie2025mhc,MGR},
and optimized multi-stream execution~\citep{zhang2026xhc}.
SimpleHC further reduces repeated residual-state access across sublayers
by sharing gate inputs while retaining layer-specific parameters.
Together with the additive residual structure, this permits consolidated
reads and deferred writes, with scalar corrections preserving sequential
output contributions for the chosen gates.

\Needspace{12\baselineskip}
\section{Conclusion}

\label{sec:conclusion}

In this work, we study how the components of the Hyper-Connections architecture
contribute to model performance and training cost.
Our study identifies learned read/write gating as a minimal effective
structure, while unconstrained mixing offers gains with propagation risks.
Beyond removing mixing, we introduce shared memory access to reduce
repeated residual-state accesses, with gate analyses supporting
layer-specific mappings conditioned on shared inputs.
Together, they form SimpleHC, simplifying residual structure and execution.
SimpleHC matches mHC performance on 15B and 30B MoE models, reduces
additional training overhead from $7.0\%$ to $2.4\%$ at the largest
tested scale, and exhibits smoother gradient dynamics while remaining
stable under stress.
SimpleHC thus combines architectural simplicity, execution efficiency,
training stability, and scalability.

%
%
%
%
%
%
\clearpage
\bibliography{references}

\begin{thebibliography}{51}
\providecommand{\natexlab}[1]{#1}
\providecommand{\url}[1]{\texttt{#1}}
\expandafter\ifx\csname urlstyle\endcsname\relax
  \providecommand{\doi}[1]{doi: #1}\else
  \providecommand{\doi}{doi: \begingroup \urlstyle{rm}\Url}\fi

\bibitem[Alimaskina et~al.(2026)Alimaskina, Molodtsov, and Beznosikov]{stream_collapse}
Ekaterina Alimaskina, Gleb Molodtsov, and Aleksandr Beznosikov.
\newblock Analyzing stream collapse in hyper-connections: From diagnosis to mitigation.
\newblock \emph{arXiv preprint arXiv:2606.03483}, 2026.

\bibitem[Austin et~al.(2021)Austin, Odena, Nye, Bosma, Michalewski, Dohan, Jiang, Cai, Terry, Le, et~al.]{MBPP}
Jacob Austin, Augustus Odena, Maxwell Nye, Maarten Bosma, Henryk Michalewski, David Dohan, Ellen Jiang, Carrie Cai, Michael Terry, Quoc Le, et~al.
\newblock Program synthesis with large language models.
\newblock \emph{arXiv preprint arXiv:2108.07732}, 2021.

\bibitem[Chen et~al.(2021)Chen, Tworek, Jun, Yuan, Pinto, Kaplan, Edwards, Burda, Joseph, Brockman, et~al.]{HumanEval}
Mark Chen, Jerry Tworek, Heewoo Jun, Qiming Yuan, Henrique Ponde De~Oliveira Pinto, Jared Kaplan, Harri Edwards, Yuri Burda, Nicholas Joseph, Greg Brockman, et~al.
\newblock Evaluating large language models trained on code.
\newblock \emph{arXiv preprint arXiv:2107.03374}, 2021.

\bibitem[Cobbe et~al.(2021)Cobbe, Kosaraju, Bavarian, Chen, Jun, Kaiser, Plappert, Tworek, Hilton, Nakano, et~al.]{GSM8k}
Karl Cobbe, Vineet Kosaraju, Mohammad Bavarian, Mark Chen, Heewoo Jun, Lukasz Kaiser, Matthias Plappert, Jerry Tworek, Jacob Hilton, Reiichiro Nakano, et~al.
\newblock Training verifiers to solve math word problems.
\newblock \emph{arXiv preprint arXiv:2110.14168}, 2021.

\bibitem[Dai et~al.(2024)Dai, Deng, Zhao, Xu, Gao, Chen, Li, Zeng, Yu, Wu, et~al.]{dai2024deepseekmoe}
Damai Dai, Chengqi Deng, Chenggang Zhao, RX~Xu, Huazuo Gao, Deli Chen, Jiashi Li, Wangding Zeng, Xingkai Yu, Yu~Wu, et~al.
\newblock Deepseekmoe: Towards ultimate expert specialization in mixture-of-experts language models.
\newblock In \emph{Proceedings of the 62nd annual meeting of the association for computational linguistics (volume 1: Long papers)}, pages 1280--1297, 2024.

\bibitem[Dandachi and Diggs-Galligan(2026)]{dandachi2026gomHC}
Torque Dandachi and Sophia Diggs-Galligan.
\newblock go-$ m $ hc: Direct parameterization of manifold-constrained hyper-connections via generalized orthostochastic matrices.
\newblock \emph{arXiv preprint arXiv:2604.02309}, 2026.

\bibitem[{Dots Studio}(2026)]{dots2026dots3note}
{Dots Studio}.
\newblock {dots3-note Preview}, 2026.
\newblock \url{https://huggingface.co/dots-studio/dots3-note-prev}.
\newblock Model card.

\bibitem[Dua et~al.(2019)Dua, Wang, Dasigi, Stanovsky, Singh, and Gardner]{dua2019drop}
Dheeru Dua, Yizhong Wang, Pradeep Dasigi, Gabriel Stanovsky, Sameer Singh, and Matt Gardner.
\newblock Drop: A reading comprehension benchmark requiring discrete reasoning over paragraphs.
\newblock In \emph{Proceedings of the 2019 Conference of the North American Chapter of the Association for Computational Linguistics: Human Language Technologies, Volume 1 (Long and Short Papers)}, pages 2368--2378, 2019.

\bibitem[Gao et~al.(2020)Gao, Biderman, Black, Golding, Hoppe, Foster, Phang, He, Thite, Nabeshima, et~al.]{gao2020pile}
Leo Gao, Stella Biderman, Sid Black, Laurence Golding, Travis Hoppe, Charles Foster, Jason Phang, Horace He, Anish Thite, Noa Nabeshima, et~al.
\newblock The pile: An 800gb dataset of diverse text for language modeling.
\newblock \emph{arXiv preprint arXiv:2101.00027}, 2020.

\bibitem[Gema et~al.(2025)Gema, Leang, Hong, Devoto, Mancino, Saxena, He, Zhao, Du, Madani, et~al.]{MMLURedux}
Aryo~Pradipta Gema, Joshua Ong~Jun Leang, Giwon Hong, Alessio Devoto, Alberto Carlo~Maria Mancino, Rohit Saxena, Xuanli He, Yu~Zhao, Xiaotang Du, Mohammad Reza~Ghasemi Madani, et~al.
\newblock Are we done with mmlu?
\newblock In \emph{Proceedings of the 2025 Conference of the Nations of the Americas Chapter of the Association for Computational Linguistics: Human Language Technologies (Volume 1: Long Papers)}, pages 5069--5096, 2025.

\bibitem[Grattafiori et~al.(2024)Grattafiori, Dubey, Jauhri, Pandey, Kadian, Al-Dahle, Letman, Mathur, Schelten, Vaughan, et~al.]{grattafiori2024llama}
Aaron Grattafiori, Abhimanyu Dubey, Abhinav Jauhri, Abhinav Pandey, Abhishek Kadian, Ahmad Al-Dahle, Aiesha Letman, Akhil Mathur, Alan Schelten, Alex Vaughan, et~al.
\newblock The llama 3 herd of models.
\newblock \emph{arXiv preprint arXiv:2407.21783}, 2024.

\bibitem[Guo et~al.(2026)Guo, Chen, Ke, Lei, Xu, Feng, Luo, et~al.]{guo2026ohc}
Haoqiang Guo, Xuyi Chen, Bo~Ke, Yishu Lei, Ziyang Xu, Shikun Feng, Wenhan Luo, et~al.
\newblock ohc: Orthogonal hyper-connections on so (4) via quaternions.
\newblock \emph{arXiv preprint arXiv:2609.02672}, 2026.

\bibitem[He et~al.(2016{\natexlab{a}})He, Zhang, Ren, and Sun]{he2016deep}
Kaiming He, Xiangyu Zhang, Shaoqing Ren, and Jian Sun.
\newblock Deep residual learning for image recognition.
\newblock In \emph{Proceedings of the IEEE conference on computer vision and pattern recognition}, pages 770--778, 2016{\natexlab{a}}.

\bibitem[He et~al.(2016{\natexlab{b}})He, Zhang, Ren, and Sun]{he2016identity}
Kaiming He, Xiangyu Zhang, Shaoqing Ren, and Jian Sun.
\newblock Identity mappings in deep residual networks.
\newblock In \emph{European conference on computer vision}, pages 630--645. Springer, 2016{\natexlab{b}}.

\bibitem[Hendrycks et~al.(2020)Hendrycks, Burns, Basart, Zou, Mazeika, Song, and Steinhardt]{MMLU}
Dan Hendrycks, Collin Burns, Steven Basart, Andy Zou, Mantas Mazeika, Dawn Song, and Jacob Steinhardt.
\newblock Measuring massive multitask language understanding.
\newblock \emph{arXiv preprint arXiv:2009.03300}, 2020.

\bibitem[Hendrycks et~al.(2021)Hendrycks, Burns, Kadavath, Arora, Basart, Tang, Song, and Steinhardt]{MATH}
Dan Hendrycks, Collin Burns, Saurav Kadavath, Akul Arora, Steven Basart, Eric Tang, Dawn Song, and Jacob Steinhardt.
\newblock Measuring mathematical problem solving with the math dataset.
\newblock \emph{arXiv preprint arXiv:2103.03874}, 2021.

\bibitem[Hu et~al.(2024)Hu, Tu, Han, He, Cui, Long, Zheng, Fang, Huang, Zhao, et~al.]{WSD}
Shengding Hu, Yuge Tu, Xu~Han, Chaoqun He, Ganqu Cui, Xiang Long, Zhi Zheng, Yewei Fang, Yuxiang Huang, Weilin Zhao, et~al.
\newblock Minicpm: Unveiling the potential of small language models with scalable training strategies.
\newblock \emph{arXiv preprint arXiv:2404.06395}, 2024.

\bibitem[Huang et~al.(2023)Huang, Bai, Zhu, Zhang, Zhang, Su, Liu, Lv, Zhang, Fu, et~al.]{ceval}
Yuzhen Huang, Yuzhuo Bai, Zhihao Zhu, Junlei Zhang, Jinghan Zhang, Tangjun Su, Junteng Liu, Chuancheng Lv, Yikai Zhang, Yao Fu, et~al.
\newblock C-eval: A multi-level multi-discipline chinese evaluation suite for foundation models.
\newblock \emph{Advances in neural information processing systems}, 36:\penalty0 62991--63010, 2023.

\bibitem[Joshi et~al.(2017)Joshi, Choi, Weld, and Zettlemoyer]{joshi2017triviaqa}
Mandar Joshi, Eunsol Choi, Daniel~S Weld, and Luke Zettlemoyer.
\newblock Triviaqa: A large scale distantly supervised challenge dataset for reading comprehension.
\newblock In \emph{Proceedings of the 55th Annual Meeting of the Association for Computational Linguistics (Volume 1: Long Papers)}, pages 1601--1611, 2017.

\bibitem[Li et~al.(2024)Li, Zhang, Koto, Yang, Zhao, Gong, Duan, and Baldwin]{li2024cmmlu}
Haonan Li, Yixuan Zhang, Fajri Koto, Yifei Yang, Hai Zhao, Yeyun Gong, Nan Duan, and Timothy Baldwin.
\newblock Cmmlu: Measuring massive multitask language understanding in chinese.
\newblock In \emph{Findings of the Association for Computational Linguistics: ACL 2024}, pages 11260--11285, 2024.

\bibitem[Liu et~al.(2024{\natexlab{a}})Liu, Feng, Wang, Wang, Liu, Zhao, Dengr, Ruan, Dai, Guo, et~al.]{liu2024deepseekV2_MLA}
Aixin Liu, Bei Feng, Bin Wang, Bingxuan Wang, Bo~Liu, Chenggang Zhao, Chengqi Dengr, Chong Ruan, Damai Dai, Daya Guo, et~al.
\newblock Deepseek-v2: A strong, economical, and efficient mixture-of-experts language model.
\newblock \emph{arXiv preprint arXiv:2405.04434}, 2024{\natexlab{a}}.

\bibitem[Liu et~al.(2024{\natexlab{b}})Liu, Feng, Xue, Wang, Wu, Lu, Zhao, Deng, Zhang, Ruan, et~al.]{liu2024deepseekv3}
Aixin Liu, Bei Feng, Bing Xue, Bingxuan Wang, Bochao Wu, Chengda Lu, Chenggang Zhao, Chengqi Deng, Chenyu Zhang, Chong Ruan, et~al.
\newblock Deepseek-v3 technical report.
\newblock \emph{arXiv preprint arXiv:2412.19437}, 2024{\natexlab{b}}.

\bibitem[Liu et~al.(2025)Liu, Su, Yao, Jiang, Lai, Du, Qin, Xu, Lu, Yan, et~al.]{liu2025muon}
Jingyuan Liu, Jianlin Su, Xingcheng Yao, Zhejun Jiang, Guokun Lai, Yulun Du, Yidao Qin, Weixin Xu, Enzhe Lu, Junjie Yan, et~al.
\newblock Muon is scalable for llm training.
\newblock \emph{arXiv preprint arXiv:2502.16982}, 2025.

\bibitem[Liu et~al.(2026)Liu, Zhang, and Li]{liu2026sHC}
Zhaoyi Liu, Haichuan Zhang, and Ang Li.
\newblock Beyond the birkhoff polytope: Spectral-sphere-constrained hyper-connections.
\newblock \emph{arXiv preprint arXiv:2603.20896}, 2026.

\bibitem[Loshchilov and Hutter(2017)]{adam}
Ilya Loshchilov and Frank Hutter.
\newblock Decoupled weight decay regularization.
\newblock \emph{arXiv preprint arXiv:1711.05101}, 2017.

\bibitem[Oldenburg et~al.(2026)Oldenburg, de~Kam, Zuijdam, Eberson, van Zutphen, de~Wildt, and Verhoeven]{mhcpeft}
Valentijn Oldenburg, Floris de~Kam, Bente Zuijdam, Lieve Eberson, Nicky van Zutphen, Stef de~Wildt, and Ivo Verhoeven.
\newblock Manifold-constrained hyper-connections for parameter-efficient finetuning.
\newblock \emph{arXiv preprint arXiv:2607.18130}, 2026.

\bibitem[Pagliardini et~al.(2024)Pagliardini, Mohtashami, Fleuret, and Jaggi]{pagliardini2024denseformer}
Matteo Pagliardini, Amirkeivan Mohtashami, Francois Fleuret, and Martin Jaggi.
\newblock Denseformer: Enhancing information flow in transformers via depth weighted averaging.
\newblock \emph{Advances in neural information processing systems}, 37:\penalty0 136479--136508, 2024.

\bibitem[Peng et~al.(2026)Peng, Rai, Tseng, Wang, and Wu]{peng2026ablateandrescue}
William Peng, Josheev Rai, Kevin Tseng, Siwei Wang, and Sean Wu.
\newblock Ablate and rescue: A causal analysis of residual stream hyper-connections.
\newblock \emph{arXiv preprint arXiv:2603.14833}, 2026.

\bibitem[Qin et~al.(2026)Qin, Zhang, Dai, Zou, Wu, Yang, Xia, Yao, Liu, and Yan]{qin2026multi}
Xiaohan Qin, Xiangdong Zhang, Tuo Dai, Sunan Zou, Huaijin Wu, Yebin Yang, Zhuo Xia, Lin Yao, Yuliang Liu, and Junchi Yan.
\newblock Multi-stream attention residuals.
\newblock \emph{arXiv preprint arXiv:2610.13630}, 2026.

\bibitem[Qiu et~al.(2026{\natexlab{a}})Qiu, Wang, Li, Li, Xu, Wang, Zhang, Men, Mao, Zhang, et~al.]{qiu2026qwen3.8}
Zihan Qiu, Zekun Wang, Xiao Li, Yanpeng Li, Yang Xu, Yixuan Wang, Huaqing Zhang, Rui Men, Bochao Mao, Chengruidong Zhang, et~al.
\newblock On the design of qwen3. 8-next architecture: Evaluation, efficiency, and training stability.
\newblock \emph{arXiv preprint arXiv:2608.30320}, 2026{\natexlab{a}}.

\bibitem[Qiu et~al.(2026{\natexlab{b}})Qiu, Wang, Zheng, Huang, Wen, Yang, Men, Yu, Huang, Huang, et~al.]{qiu2026gated}
Zihan Qiu, Zekun Wang, Bo~Zheng, Zeyu Huang, Kaiyue Wen, Songlin Yang, Rui Men, Le~Yu, Fei Huang, Suozhi Huang, et~al.
\newblock Gated attention for large language models: Non-linearity, sparsity, and attention-sink-free.
\newblock \emph{Advances in Neural Information Processing Systems}, 38:\penalty0 100092--100118, 2026{\natexlab{b}}.

\bibitem[Sun et~al.(2020)Sun, Yu, Yu, and Cardie]{C3}
Kai Sun, Dian Yu, Dong Yu, and Claire Cardie.
\newblock Investigating prior knowledge for challenging chinese machine reading comprehension.
\newblock \emph{Transactions of the Association for Computational Linguistics}, 8:\penalty0 141--155, 2020.

\bibitem[Suzgun et~al.(2023)Suzgun, Scales, Sch{\"a}rli, Gehrmann, Tay, Chung, Chowdhery, Le, Chi, Zhou, et~al.]{BBH}
Mirac Suzgun, Nathan Scales, Nathanael Sch{\"a}rli, Sebastian Gehrmann, Yi~Tay, Hyung~Won Chung, Aakanksha Chowdhery, Quoc Le, Ed~H Chi, Denny Zhou, et~al.
\newblock Challenging big-bench tasks and whether chain-of-thought can solve them.
\newblock In \emph{Findings of the Association for Computational Linguistics: ACL 2023}, pages 13003--13051, 2023.

\bibitem[Team et~al.(2026)Team, Chen, Zhang, Su, Xu, Pan, Wang, Wang, Chen, Yin, et~al.]{team2026attention_residual}
Kimi Team, Guangyu Chen, Yu~Zhang, Jianlin Su, Weixin Xu, Siyuan Pan, Yaoyu Wang, Yucheng Wang, Guanduo Chen, Bohong Yin, et~al.
\newblock Attention residuals.
\newblock \emph{arXiv preprint arXiv:2603.15031}, 2026.

\bibitem[{Tencent Hy Team}(2026)]{tencent2026hy4preview}
{Tencent Hy Team}.
\newblock {Hy4 preview}, 2026.
\newblock \url{https://huggingface.co/tencent/Hy4-preview}.
\newblock Model card.

\bibitem[Tillet et~al.(2019)Tillet, Kung, and Cox]{tillet2019triton}
Philippe Tillet, H.~T. Kung, and David Cox.
\newblock {Triton}: An intermediate language and compiler for tiled neural network computations.
\newblock In \emph{Proceedings of the 3rd ACM SIGPLAN International Workshop on Machine Learning and Programming Languages}. Association for Computing Machinery, 2019.
\newblock \doi{10.1145/3315508.3329973}.
\newblock \url{https://doi.org/10.1145/3315508.3329973}.

\bibitem[Vaswani et~al.(2017)Vaswani, Shazeer, Parmar, Uszkoreit, Jones, Gomez, Kaiser, and Polosukhin]{vaswani2017attention}
Ashish Vaswani, Noam Shazeer, Niki Parmar, Jakob Uszkoreit, Llion Jones, Aidan~N Gomez, {\L}ukasz Kaiser, and Illia Polosukhin.
\newblock Attention is all you need.
\newblock \emph{Advances in neural information processing systems}, 30, 2017.

\bibitem[Wang et~al.(2024)Wang, Ma, Zhang, Ni, Chandra, Guo, Ren, Arulraj, He, Jiang, et~al.]{wang2024mmlupro}
Yubo Wang, Xueguang Ma, Ge~Zhang, Yuansheng Ni, Abhranil Chandra, Shiguang Guo, Weiming Ren, Aaran Arulraj, Xuan He, Ziyan Jiang, et~al.
\newblock Mmlu-pro: A more robust and challenging multi-task language understanding benchmark.
\newblock \emph{Advances in Neural Information Processing Systems}, 37:\penalty0 95266--95290, 2024.

\bibitem[Wortsman et~al.(2024)Wortsman, Liu, Xiao, Everett, Alemi, Adlam, Co-Reyes, Gur, Kumar, Novak, et~al.]{wortsman2024small}
Mitchell Wortsman, Peter Liu, Lechao Xiao, Katie Everett, Alexander Alemi, Ben Adlam, John~D Co-Reyes, Izzeddin Gur, Abhishek Kumar, Roman Novak, et~al.
\newblock Small-scale proxies for large-scale transformer training instabilities.
\newblock In \emph{International Conference on Learning Representations}, volume 2024, pages 49844--49869, 2024.

\bibitem[Xiao et~al.(2025)Xiao, Meng, Li, and Yuan]{xiao2025muddformer}
Da~Xiao, Qingye Meng, Shengping Li, and Xingyuan Yuan.
\newblock Muddformer: Breaking residual bottlenecks in transformers via multiway dynamic dense connections.
\newblock \emph{arXiv preprint arXiv:2502.12170}, 2025.

\bibitem[Xie et~al.(2025)Xie, Wei, Cao, Zhao, Deng, Li, Dai, Gao, Chang, Yu, et~al.]{xie2025mhc}
Zhenda Xie, Yixuan Wei, Huanqi Cao, Chenggang Zhao, Chengqi Deng, Jiashi Li, Damai Dai, Huazuo Gao, Jiang Chang, Kuai Yu, et~al.
\newblock mhc: Manifold-constrained hyper-connections.
\newblock \emph{arXiv preprint arXiv:2512.24880}, 2025.

\bibitem[Xu et~al.(2026)Xu, Lin, Xue, Wang, Xu, Wu, Zhang, Lin, Dong, Ling, et~al.]{xu2026deepseek}
Anyi Xu, Bangcai Lin, Bing Xue, Bingxuan Wang, Bingzheng Xu, Bochao Wu, Bowei Zhang, Chaofan Lin, Chen Dong, Chenchen Ling, et~al.
\newblock Deepseek-v4: Towards highly efficient million-token context intelligence.
\newblock \emph{arXiv preprint arXiv:2606.19348}, 2026.

\bibitem[Yang et~al.(2025)Yang, Li, Yang, Zhang, Hui, Zheng, Yu, Gao, Huang, Lv, et~al.]{yang2025qwen3}
An~Yang, Anfeng Li, Baosong Yang, Beichen Zhang, Binyuan Hui, Bo~Zheng, Bowen Yu, Chang Gao, Chengen Huang, Chenxu Lv, et~al.
\newblock Qwen3 technical report.
\newblock \emph{arXiv preprint arXiv:2505.09388}, 2025.

\bibitem[Yang and Gao(2026)]{yang2026mhclite}
Yongyi Yang and Jianyang Gao.
\newblock mhc-lite: You don't need 20 sinkhorn-knopp iterations.
\newblock \emph{arXiv preprint arXiv:2601.05732}, 2026.

\bibitem[Zellers et~al.(2019)Zellers, Holtzman, Bisk, Farhadi, and Choi]{zellers2019hellaswag}
Rowan Zellers, Ari Holtzman, Yonatan Bisk, Ali Farhadi, and Yejin Choi.
\newblock Hellaswag: Can a machine really finish your sentence?
\newblock In \emph{Proceedings of the 57th annual meeting of the association for computational linguistics}, pages 4791--4800, 2019.

\bibitem[Zeng et~al.(2026)Zeng, Lv, Hou, Du, Zheng, Chen, Yin, Ge, Huang, Xie, et~al.]{GLM53flash}
Aohan Zeng, Xin Lv, Zhenyu Hou, Zhengxiao Du, Qinkai Zheng, Bin Chen, Da~Yin, Chendi Ge, Chenghua Huang, Chengxing Xie, et~al.
\newblock Glm-5: from vibe coding to agentic engineering.
\newblock \emph{arXiv preprint arXiv:2602.15763}, 2026.

\bibitem[Zhang et~al.(2026)Zhang, Qin, Zou, Dai, Shi, Wu, Yang, Xia, Zhang, Yao, et~al.]{zhang2026xhc}
Xiangdong Zhang, Xiaohan Qin, Sunan Zou, Tuo Dai, Xiaoming Shi, Huaijin Wu, Yebin Yang, Zhuo Xia, Shaofeng Zhang, Lin Yao, et~al.
\newblock xhc: Expanded hyper-connections.
\newblock \emph{arXiv preprint arXiv:2607.14530}, 2026.

\bibitem[Zhao et~al.(2025)Zhao, Xu, Zhao, Li, Xu, Xu, Liu, Zhou, and Yu]{deepgemm2025}
Chenggang Zhao, Zhean Xu, Liang Zhao, Jiashi Li, Chenhao Xu, Anyi Xu, Shengyu Liu, Kexing Zhou, and Kuai Yu.
\newblock {DeepGEMM}: clean and efficient {BLAS} kernel library on {GPU}.
\newblock \url{https://github.com/deepseek-ai/DeepGEMM}, 2025.

\bibitem[Zheng et~al.(2026)Zheng, Zhang, Liu, Xia, Liu, Hu, and Zhou]{MGR}
Zhizhan Zheng, Feiyun Zhang, Shuchun Liu, Tian Xia, Xi~Liu, Dasheng Hu, and Hongquan Zhou.
\newblock Multi-gate residuals.
\newblock \emph{arXiv preprint arXiv:2605.23259}, 2026.

\bibitem[Zhou et~al.(2026)Zhou, Gu, Iacovides, and Mandic]{zhou2026kromhc}
Wuyang Zhou, Yuxuan Gu, Giorgos Iacovides, and Danilo Mandic.
\newblock Kromhc: Manifold-constrained hyper-connections with kronecker-product residual matrices.
\newblock \emph{arXiv preprint arXiv:2601.21579}, 2026.

\bibitem[Zhu et~al.(2025)Zhu, Huang, Huang, Zeng, Mao, Wu, Min, and Zhou]{zhu2025hyper}
Defa Zhu, Hongzhi Huang, Zihao Huang, Yutao Zeng, Yunyao Mao, Banggu Wu, Qiyang Min, and Xun Zhou.
\newblock Hyper-connections.
\newblock In \emph{International Conference on Learning Representations}, volume 2025, pages 97183--97219, 2025.

\end{thebibliography}
\bibliographystyle{assets/plainnat}

\clearpage
\appendix
\section*{Appendix}

\section{Experimental Settings}
\label{app:experimental-details}

This section provides additional details on the experimental settings. Tables~\ref{tab:map} and~\ref{tab:share} use the 15B-A1B setup,
and Table~\ref{tab:flagship_results} uses the 30B-A2B setup.
Efficiency measurements use the separate configurations in
Appendix~\ref{app:efficiency-settings}. Gate-analysis and stability figures
are documented in Appendices~\ref{app:gate-analysis}
and~\ref{app:stability-details}, respectively.

\subsection{Model and Training Configurations}
\label{app:training-config}

Table~\ref{tab:training-config} reports the backbone architectures and
training recipes for the 15B-A1B design-space studies and the 30B-A2B
scaling experiments. Parameter counts are rounded; activated counts exclude embedding
and output layers.
Both backbones apply headwise sigmoid gating to attention outputs before
the output projection~\citep{qiu2026gated}.

\begin{table}[!t]
\centering
\small
\setlength{\tabcolsep}{10pt}
\renewcommand{\arraystretch}{1.08}
\caption{\textbf{Model configurations and training hyperparameters.}
Residual streams apply to HC variants, and group size applies to SimpleHC,
unless varied in the corresponding experiments.}
\label{tab:training-config}
\vspace{5pt}
\begin{tabular}{l|cc}
\toprule
\multicolumn{1}{l}{\textbf{Attribute}} & \textbf{15B-A1B} & \textbf{30B-A2B} \\
\midrule
Total parameters & 15B & 30B \\
Activated parameters & 1B & 2B \\
\midrule
Transformer layers & 24 & 32 \\
Hidden dimension & 1792 & 2048 \\
Vocabulary size & 152,064 & 152,064 \\
Attention & MLA & MLA \\
Full / sliding-window ratio & 1:3 & 1:3 \\
Sliding-window size & 512 & 512 \\
Attention heads & 40 & 56 \\
Experts (routed / active / shared) & 256 / 8 / 1 & 256 / 8 / 1 \\
Expert intermediate dimension & 448 & 576 \\
\midrule
Residual streams $n$ & 4 & 4 \\
Group size $K$ (SimpleHC) & 4 & 4 \\
\midrule
Sequence length & 8192 & 8192 \\
Global batch size & - & 768 \\
Training tokens & - & - \\
Optimizer & Muon / AdamW & Muon / AdamW \\
Peak learning rate & $9\times10^{-4}$ & $7.4\times10^{-4}$ \\
Learning-rate schedule & WSD, arccos decay & WSD, arccos decay \\
Warmup steps & 1,000 & 1,000 \\
Decay fraction & $20\%$ & $20\%$ \\
Weight decay & 0.1 & 0.1 \\
Gradient clipping & 1.0 & 1.0 \\
Training precision & BF16 & BF16 \\
\bottomrule
\end{tabular}

\end{table}

\textbf{Training hyperparameters.}
Table~\ref{tab:training-config} summarizes the model configurations and
training hyperparameters. Muon updates backbone matrix weights in attention and MLP/MoE modules;
AdamW updates embeddings, output weights, normalization parameters,
and HC gate parameters.
We use a warmup--stable--decay (WSD)~\citep{WSD} learning-rate schedule with 1,000
warmup steps, followed by a constant learning rate and arccos decay over
the final $20\%$ of training.

For the gate-reuse variant in Table~\ref{tab:share}, each group
uses one learned read/write gate pair instead of four layer-specific pairs.
The gates are computed from the group-entry state and reused across all sublayers.

\textbf{Initialization.}
We follow the initialization scheme of the mHC implementation in the
xHC repository.\footnote{\label{fn:xhc-repo}\url{https://github.com/aHapBean/xHC}}
For identityHC and SimpleHC, we use an analogous initialization scheme
for the read/write gate parameters, with residual transitions fixed to
the identity.

\subsection{Efficiency Benchmark Settings}
\label{app:efficiency-settings}
We evaluate training efficiency across model scales by varying hidden width
from 1024 to 6144, as summarized in Table~\ref{tab:efficiency-config}.
To limit memory requirements and avoid introducing parallelism
changes, all benchmark models use eight Transformer layers. This allows
comparisons across widths without the resource demands of full-depth models.
At each width, residual variants share the backbone, batch sizes, and
parallel configuration. Since both residual operations and backbone computation repeat across
layers, these measurements provide a representative estimate of relative
training overhead in deeper models with the same parallel configuration.

All models use MLA with one full-attention layer followed by three
sliding-window layers of window size 512. The first layer has a dense FFN;
the remaining seven use 256 routed experts with top-8 routing and one
shared expert. We use BF16, a sequence length of 8192, and a global batch
size of 256. CUDA Graph is enabled for all methods, with activation
recomputation disabled. mHC, identityHC, and SimpleHC use four residual
streams, and SimpleHC groups four sublayers.

\textbf{xHC comparisons.}
We benchmark xHC under the same timing protocol described
above (Figure~\ref{fig:efficiency-xhc}). Our efficient xHC implementation
is numerically validated against the official implementation.\textsuperscript{\ref{fn:xhc-repo}}
xHC uses $N=16$ residual streams with $k=4$ active streams.

\begin{table}[!t]
\centering
\small
\setlength{\tabcolsep}{8pt}
\renewcommand{\arraystretch}{1.08}
\caption{\textbf{Model configurations for efficiency measurements.}
Each row defines an eight-layer model. FFN widths denote intermediate
dimensions; the shared expert uses the same width as each routed expert.
Microbatch sizes count sequences and are shared across methods at each width.}
\label{tab:efficiency-config}
\begin{tabular}{r|rrrrr}
\toprule
Width & Heads & Query rank & Dense FFN & Expert FFN & Microbatch \\
\midrule
1024 & 24 & 256 & 2880 & 320 & 8 \\
2048 & 48 & 384 & 5728 & 608 & 4 \\
3072 & 80 & 640 & 8608 & 928 & 2 \\
4096 & 96 & 768 & 11456 & 1216 & 1 \\
5120 & 128 & 1024 & 14336 & 1536 & 1 \\
6144 & 160 & 1280 & 17216 & 1856 & 1 \\
\bottomrule
\end{tabular}

\end{table}

\textbf{Saved activation storage accounting.}
\label{app:saved-activation-accounting}
We estimate storage retained for backward after one microbatch forward
pass, covering sublayer inputs and outputs, residual states and copies,
base reads, HC auxiliary tensors, and the decoder output before final
normalization (Table~\ref{tab:saved-activation-breakdown}). Shared storage
is counted once, while distinct copies are counted separately. The
accounting excludes attention/FFN internals, LM-head/loss buffers, model
and optimizer states, temporary workspaces, and allocator/CUDA Graph
reservations. These source-level estimates are not total activation or
peak device memory. Figure~\ref{fig:simplehc-overview}(b) uses microbatch
size one; both estimates scale linearly with microbatch size, leaving
the relative reduction unchanged.

The models use eight layers (16 sublayers), $n=4$, and no recomputation;
SimpleHC uses four groups with $K=4$. For $T$ local microbatch tokens,
let $U=2TC$ bytes denote one BF16 narrow activation. The mHC residual
inputs occupy $U+15\times4U=61U$: the first uses a broadcast view of
the dense input, followed by 15 wide inputs. SimpleHC retains two wide
copies for gate projection and base reads in the first group, and one
per later group, totaling $8U+3\times4U=20U$. Each group jointly allocates
four base reads; the first sublayer input keeps the entire allocation
live through a view. Beyond that input, already counted in the input
row, the remaining slices contribute $3U$ per group, or $12U$ overall.
identityHC has a similar saved-storage footprint to mHC, as both retain wide residual states at each sublayer.

\begin{table}[!t]
\centering
\small
\setlength{\tabcolsep}{5pt}
\renewcommand{\arraystretch}{1.08}
\caption{\textbf{Estimated activation storage retained for backward
at sublayer boundaries and within HC connections.}
The example uses eight Transformer layers, $C=2048$, and one
8192-token sequence; $1\,\mathrm{MiB}=2^{20}$ bytes.
Shared storage is counted once, including allocations retained by views.}
\label{tab:saved-activation-breakdown}
\begin{tabular}{lrrrr}
\toprule
& \multicolumn{2}{c}{Storage (bytes)} & \multicolumn{2}{c}{Example (MiB)} \\
\cmidrule(lr){2-3}\cmidrule(lr){4-5}
Component & mHC & SimpleHC & mHC & SimpleHC \\
\midrule
Pre-normalization sublayer inputs & $16U$ & $16U$ & 512 & 512 \\
Residual states and input copies & $61U$ & $20U$ & 1952 & 640 \\
Sublayer branch outputs & $16U$ & $16U$ & 512 & 512 \\
Additional base-read storage & $0$ & $12U$ & 0 & 384 \\
Collapsed decoder output & $U$ & $U$ & 32 & 32 \\
HC auxiliary tensors & $4672T$ & $1648T$ & 36.5 & 12.875 \\
\midrule
Total & $94U+4672T$ & $65U+1648T$ & 3044.5 & 2092.875 \\
\bottomrule
\end{tabular}

\end{table}

HC auxiliary tensors are FP32 and include projections, gates, inverse-RMS
statistics, and mixing or correction intermediates. The mHC path retains
73 elements per token per sublayer ($4672T$ bytes overall); SimpleHC
retains 103 per token per group ($1648T$ bytes), including contiguous gate
copies and correction coefficients. At $C=2048$ and $T=8192$, total storage
falls from 3044.5 to 2092.875 MiB, a $31.26\%$ reduction. Across the six
evaluated widths, reductions range from $31.0\%$ to $31.7\%$, supporting
the rounded $31\%$ summary.

\subsection{Evaluation Protocols}
\label{app:eval-protocols}

We report validation loss on the Pile test split~\citep{gao2020pile} in
Tables~\ref{tab:map}, \ref{tab:share}, and~\ref{tab:flagship_results}.

For the 15B-A1B component studies, we evaluate MMLU~\citep{MMLU}
in a 5-shot setting, HellaSwag~\citep{zellers2019hellaswag} in a
10-shot setting, MATH~\citep{MATH} in a 4-shot setting, and
MBPP~\citep{MBPP} and C3~\citep{C3}, both in a 3-shot setting.
For the 30B-A2B evaluation, we retain these settings and add
MMLU-Pro~\citep{wang2024mmlupro}, MMLU-Redux~\citep{MMLURedux},
TriviaQA~\citep{joshi2017triviaqa}, CMMLU~\citep{li2024cmmlu},
and CEval~\citep{ceval} with 5-shot evaluation;
BBH~\citep{BBH} and DROP~\citep{dua2019drop} with 3-shot evaluation;
GSM8K~\citep{GSM8k} with 4-shot evaluation; and
HumanEval~\citep{HumanEval} with zero-shot evaluation.

\section{Exact Reconstruction with Scalar Corrections}
\label{app:execution-proof}
\label{app:shared-execution}

We analyze the forward execution in Algorithm~\ref{alg:shared-schedule}.

We prove that scalar reconstruction and deferred write-back are equivalent
to sequential execution with the same shared-input gates. Both executions
use Equation~\ref{eq:shared-input-gates}, identical parameters, and identical
random choices for any stochastic operations. Define the sequential
reference by $\widetilde X_L=X_L$ and
\begin{equation}
 \widetilde u_j=\widetilde X_j^\top\bar h_j^{\mathrm{pre}},\qquad
 \widetilde o_j=F_j(\widetilde u_j),\qquad
 \widetilde X_{j+1}=\widetilde X_j+
                   \bar h_j^{\mathrm{post}}\widetilde o_j^\top.
 \label{eq:shared-sequential-reference}
\end{equation}

\textbf{Forward equivalence.}
At $j=L$, Equation~\ref{eq:scalar-reconstruction} has no correction terms,
so $u_L=\widetilde u_L$ and $o_L=\widetilde o_L$.
Assume $o_k=\widetilde o_k$ for $L\leq k<j$. Expanding the reference
updates and applying the read decomposition in
Equation~\ref{eq:rank-one-read} gives
\begin{equation}
 \widetilde u_j
 =X_L^\top\bar h_j^{\mathrm{pre}}
  +\sum_{k=L}^{j-1}
    \bigl((\bar h_j^{\mathrm{pre}})^\top\bar h_k^{\mathrm{post}}\bigr)
    \widetilde o_k
 =u_j.
\end{equation}
Thus $\widetilde o_j=o_j$, completing the induction. Summing the reference
updates also yields exactly the group-boundary state in
Equation~\ref{eq:group-boundary-write}.

\textbf{Gradient equivalence.}
The two executions define the same function of $X_L$ and the parameters,
so their derivatives agree wherever this function is differentiable.
Gradients must include the dependence of every gate on $X_L$; detaching
gates would invalidate this equivalence. Nondifferentiable operations
require consistent derivative conventions. The equivalence holds in exact
arithmetic; floating-point results may differ with reduction order and
precision. It does not assert equivalence to gates conditioned on each
updated sublayer state.

\Needspace{8\baselineskip}
\section{Read/Write Gating and Unconstrained Mixing}
\label{app:interaction-groups}

We examine the weighting mechanisms of the two interaction groups in
Table~\ref{tab:map} to interpret their comparable empirical performance.
Both control how earlier outputs contribute to later inputs, but
place learnable flexibility at different points along the residual path.
For a fixed token, the residual-path coefficient carrying output $o_k$
into a later input $u_j$ is
\begin{equation}
 c_{j,k}=(h_j^{\mathrm{pre}})^\top
 \Phi_{k+1\to j}h_k^{\mathrm{post}},\qquad
 \Phi_{k+1\to j}=H_{j-1}^{\mathrm{res}}\cdots H_{k+1}^{\mathrm{res}},
 \label{eq:residual-contribution-coefficient}
\end{equation}
where an empty product is $I$. With learned read/write gates and identity
transitions, this reduces to
\begin{equation}
 c_{j,k}=(h_j^{\mathrm{pre}})^\top h_k^{\mathrm{post}}.
\end{equation}
For unconstrained mixing with the fixed gates used in our comparisons,
$h_j^{\mathrm{pre}}=\mathbf{1}/n$ and $h_k^{\mathrm{post}}=\mathbf{1}$,
the same coefficient becomes
\begin{equation}
 c_{j,k}=\frac{1}{n}\mathbf{1}^\top
 \Phi_{k+1\to j}\mathbf{1}.
\end{equation}

Learned read/write gates jointly determine historical-output weights:
the write gate controls how an output is allocated across streams,
while a later read gate determines how that allocation contributes
to its input. Unconstrained mixing instead modifies the allocation
through intervening residual transitions, even with fixed read/write gates.

Under the mHC gate parameterization, read and write gates lie in
$(0,1)$ and $(0,2)$, respectively, bounding their inner-product
coefficients between $0$ and $2n$. Unconstrained mixing permits
signed coefficients without a comparable parameter-independent bound.
It therefore offers additional weighting freedom, but also introduces
the propagation risks discussed in the main text.

These mechanisms provide alternative ways to learn effective
historical-output weights, offering a structural explanation for why
either group can achieve competitive performance. Learning both groups
combines these controls, consistent with the further gains in
Table~\ref{tab:map}. Their comparable scores remain an empirical finding
rather than a consequence of equivalent expressivity.

\section{Gate Analysis Details}
\label{app:gate-analysis}
\label{app:gate-sensitivity}

This section details the measurements and experimental protocols
for the analyses in Sections~\ref{sec:analysis-minimal}
and~\ref{sec:analysis-sharing}.

\subsection{Residual-Path Contribution Coefficients}
\label{app:transfer-diagnostics}

\begin{figure}[!t]
\centering
\begin{minipage}[t]{0.48\textwidth}
\centering
\includegraphics[width=\linewidth]{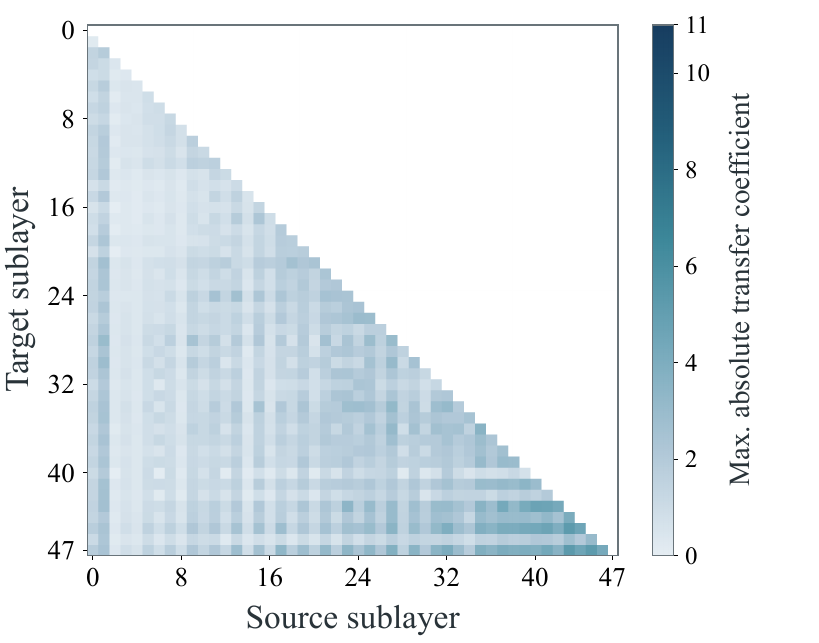}
\par
{\small (a) Learned gates with identity transitions}
\end{minipage}\hfill
\begin{minipage}[t]{0.48\textwidth}
\centering
\includegraphics[width=\linewidth]{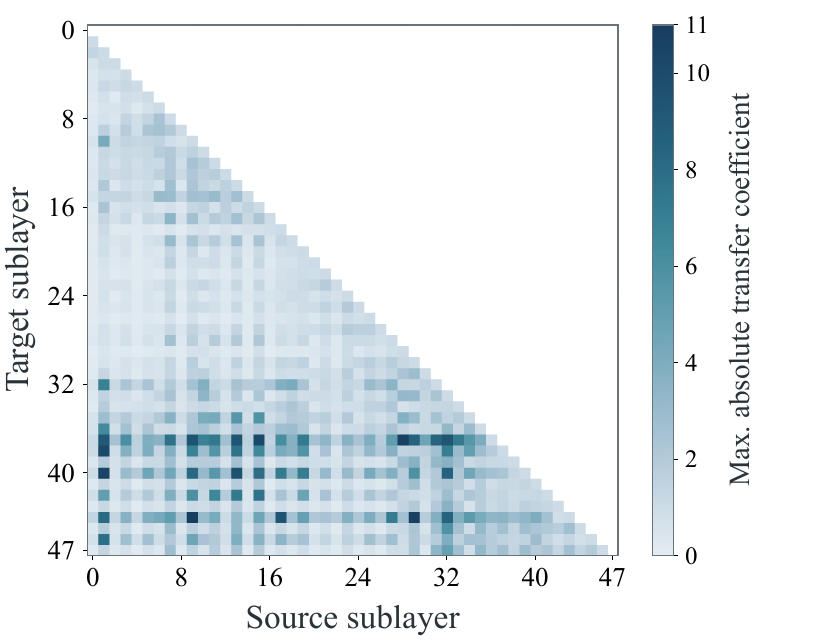}
\par
{\small (b) Unconstrained mixing with fixed gates}
\end{minipage}

\caption{\textbf{Learned read/write gating avoids the larger coefficient peaks observed with unconstrained mixing.}
Blank cells indicate undefined
source--target pairs.}
\label{fig:residual-transfer}

\end{figure}

Figure~\ref{fig:residual-transfer} compares the final checkpoints of two
15B-A1B models: unconstrained mixing with fixed read/write
gates and learned read/write gates with identity transitions. Both have
24 Transformer blocks, 48 sublayers, and four residual streams
(Table~\ref{tab:map}, rows 12 and 6).
We use 1,024 analysis sequences and sample 256 non-special-token positions
per sequence with a fixed seed. Both models use the same 262,144 token
occurrences. Sublayer indices alternate attention and FFN, from 0 to 47.

For each source $a$ and later target $b$, we evaluate the coefficient
$c_{b,a}$ in Equation~\ref{eq:residual-contribution-coefficient} and plot
$M_{b,a}=\max_{t\in\mathcal T}|c_{b,a}(t)|$.
The mixing configuration fixes $h^{\mathrm{pre}}=\mathbf{1}/4$ and
$h^{\mathrm{post}}=\mathbf{1}$; the identity configuration uses learned
gates. These coefficients use raw write gates, without division by two.
Each cell may attain its maximum at a different token. These sampled residual-path coefficients are neither matrix norms
nor full-network Jacobians and do not alone establish training instability
or increasing amplification with propagation distance.

\subsection{Gate Variation Across Sublayers}
\label{app:gate-variation}

\begin{figure}[!t]
\centering
\begin{minipage}[t]{0.48\textwidth}
\centering
\includegraphics[width=\linewidth]{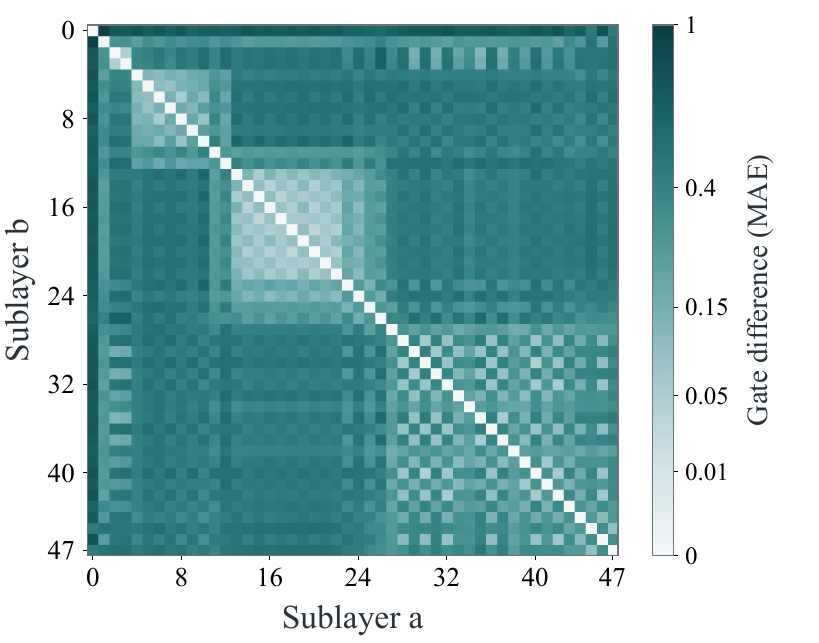}
\par
{\small (a) Read-gate variation across sublayers}
\end{minipage}\hfill
\begin{minipage}[t]{0.48\textwidth}
\centering
\includegraphics[width=\linewidth]{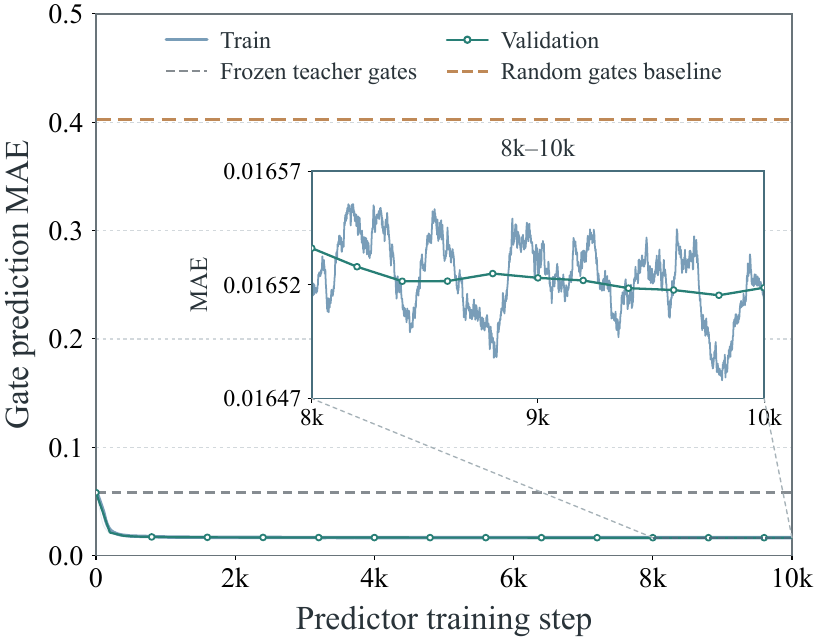}
\par
{\small (b) Prediction from the group-entry state}
\end{minipage}

\caption{\textbf{Read gates vary across sublayers yet are predictable from shared inputs.}
(a) Pairwise read-gate MAE across identityHC sublayers, averaged over
streams and matched tokens.
(b) Training and validation MAE for predicting read gates of the next
three sublayers from the group-entry state, using the same experiment
as Figure~\ref{fig:gate-sharing-analysis}(b).
Read gates are evaluated on their original $[0,1]$ scale;
the inset enlarges steps 8k--10k.}
\label{fig:read-gate-analysis}

\end{figure}

We use the same identityHC checkpoint and sampled token occurrences.
For sublayers $a,b$, the plotted difference is
\begin{equation}
 D^s(a,b)=\frac{1}{4|\mathcal T|}
 \sum_{t\in\mathcal T}\sum_{i=1}^{4}
 \left|q_{a,i}^s(t)-q_{b,i}^s(t)\right|,
 \qquad q^{\mathrm{pre}}=h^{\mathrm{pre}},\quad
 q^{\mathrm{post}}=h^{\mathrm{post}}/2.
\end{equation}
Absolute differences are computed for matched token positions and stream
indices before averaging. Figure~\ref{fig:gate-sharing-analysis}(a) shows
write gates; Figure~\ref{fig:read-gate-analysis}(a) shows read gates.
Both use a power-normalized color scale with $\gamma=0.4$ over $[0,1]$;
colorbar labels give actual MAE. The matrices are symmetric with zero
diagonals and measure gate variation rather than causal propagation.

\subsection{Prediction from the Group-Entry State}
\label{app:gate-prediction}

The prediction experiment freezes the final identityHC checkpoint and
groups its 48 sublayers into 12 groups of $K=4$. The teacher always
executes with its own gates. Predictors receive only the same-token
group-entry state $X_L$; their outputs never alter the teacher trajectory.

\textbf{Predictors and targets.}
For $L\leq j<L+K$ and $s\in\{\mathrm{pre},\mathrm{post}\}$, the target is
$h_j^s=g_j^s(X_j)$ from the frozen teacher. Each predictor uses the
SimpleHC gate parameterization with independent parameters for every
sublayer and gate type:
\begin{equation}
 \begin{aligned}
 z_L&=\operatorname{RMSNorm}(\operatorname{vec}(X_L)),\\
 \widehat\ell_j^s&=\widehat\alpha_j^s
       (\widehat W_j^s)^\top z_L+\widehat b_j^s,\\
 \widehat h_j^{\mathrm{pre}}&=\sigma(\widehat\ell_j^{\mathrm{pre}}),\qquad
 \widehat h_j^{\mathrm{post}}=2\sigma(\widehat\ell_j^{\mathrm{post}}).
 \end{aligned}
\end{equation}
The input dimension is $4\times1792=7168$, with 32 gate values per group
and approximately 2.75M trainable parameters in total. Projections, scales, and biases
are initialized from the corresponding teacher parameters. Read and write
gates are trained jointly; teacher states and targets are detached.

\textbf{Training and evaluation.}
We train the predictor for 10,000 steps with MSE loss, a global batch size
of 64, and a sequence length of 8,192. We use AdamW~\citep{adam} with
$\beta=(0.9,0.999)$, $\epsilon=10^{-8}$, zero weight decay, and gradient
clipping at 1.0. The learning rate increases
linearly from zero to $10^{-3}$ over 500 warmup steps, then follows cosine
decay to $10^{-5}$. Validation uses a fixed set of 2,048 sequences,
excluding special tokens. We evaluate every 200 steps and use the final
step-10,000 checkpoint for testing, without dev-based checkpoint selection.

\textbf{Baselines and metrics.}
The frozen reference applies each sublayer mapping copied from the teacher
to the shared entry state without adaptation. It remains input-dependent
and does not reuse one gate pair across sublayers. In the plots, this
reference is labeled ``Frozen teacher gates.'' The random baseline samples
independent $h^{\mathrm{pre}}\sim\mathcal U(0,1)$ and
$h^{\mathrm{post}}\sim\mathcal U(0,2)$ for each token, sublayer, and stream.
The constant baseline uses mean gates fitted on training data.
MAE is computed by averaging absolute errors over valid tokens, groups,
future offsets 1--3, and streams;
write gates are divided by two. Offset 0 is excluded.

\textbf{Experimental results.}
On 102,400 test sequences, the final predictor achieves read/write MAE
of 0.01653/0.01034, compared with 0.05817/0.05044 for frozen mappings
and 0.07345/0.06893 for constant mean gates.
Together with the improvements over uniform random gates on the dev set,
these results support closely approximating subsequent gates from shared
entry states using layer-specific mappings.

\section{Training Stability Details}
\label{app:stability-details}

This section details the stress-test design, training dynamics, and
instability diagnostics supporting Section~\ref{sec:disc-stability}.

\subsection{Stress-Test Design and Settings}
To emulate stability challenges in large-scale training, we combine
64 layers, 512 routed experts with top-8 routing, and a learning rate
$10\times$ the reference value. This stresses optimization through a
longer propagation path, higher sparsity, and larger updates.
A hidden width of 2816 keeps activated parameters near 6B, yielding a
230B-A6B model. The experiment tests robustness under these combined
pressures rather than isolating individual factors.

\begin{table}[!t]
\centering
\small
\setlength{\tabcolsep}{12pt}
\caption{\textbf{Model and training configuration for the stress test.}}
\label{tab:stability-config}
\begin{tabular}{lr}
\toprule
Configuration & Value \\
\midrule
Transformer layers & 64 \\
Hidden width & 2816 \\
Routed experts & 512 \\
Experts activated per token & 8 \\
Activated parameters & 6B \\
Total parameters & 230B \\
\midrule
Peak learning rate & $6.2\times10^{-3}$ \\
Global batch size & - \\
Sequence length & 8192 \\
Warmup steps & 4000 \\
\bottomrule
\end{tabular}

\end{table}

All variants use the backbone in Table~\ref{tab:stability-config}.
MLA alternates full and sliding-window attention in a 1:3 pattern,
with 72 heads and window size 512. The first FFN is dense with width
7936; the remaining 63 use experts of width 832 and one shared expert.
Training uses BF16, microbatch size 1, and Muon with momentum 0.95,
weight decay 0.1, and gradient clipping at 1.0. The WSD schedule uses
4,000 linear warmup steps, followed by the peak learning rate throughout
the comparison window. HC variants use $n=4$; mHC uses 20 Sinkhorn
iterations, and SimpleHC uses $K=4$.

\subsection{Training Dynamics}
\textbf{Standard training.}
During 15B training, SimpleHC shows smaller gradient-norm fluctuations
early after warmup than mHC and identityHC, with comparable ranges later
(Figure~\ref{fig:training-stability}(a)). The maximum 1,000-step rolling
standard deviation after warmup is 0.032 for SimpleHC, versus 0.206 for
mHC and 0.248 for identityHC, indicating smoother standard training.

\begin{figure}[!t]
\centering
\includegraphics[width=0.6\linewidth]{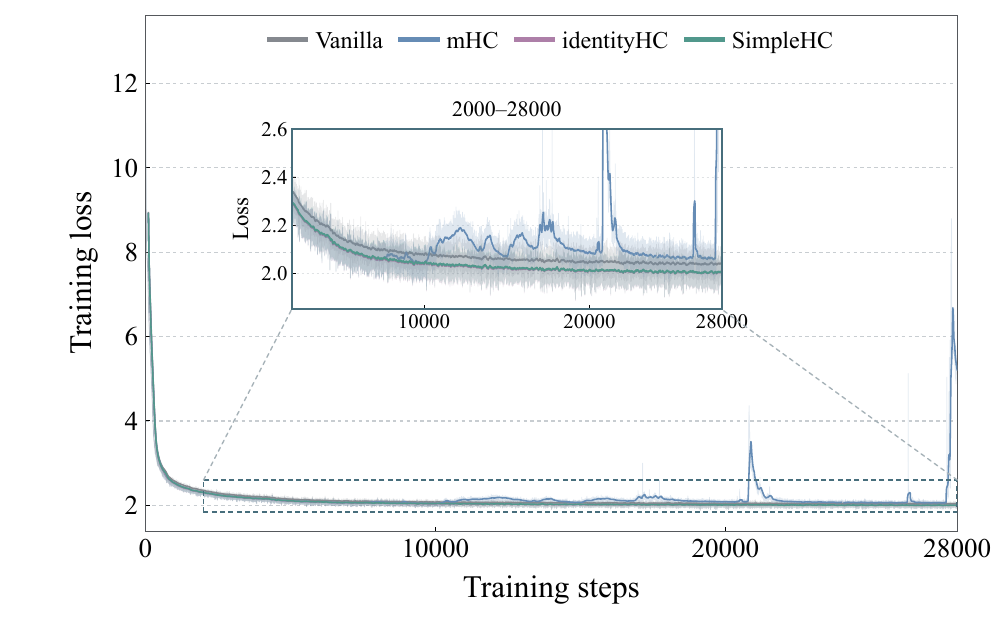}
\caption{\textbf{Training loss under stress.}
All runs cover the same 28,000 steps.
We show per-step losses and their 100-step moving averages.
The inset enlarges the low-loss region over steps 2,000--28,000.}
\label{fig:stress-loss}

\end{figure}

\begin{figure}[!t]
\centering
\begin{minipage}[t]{0.49\textwidth}
\centering
\includegraphics[width=\linewidth]{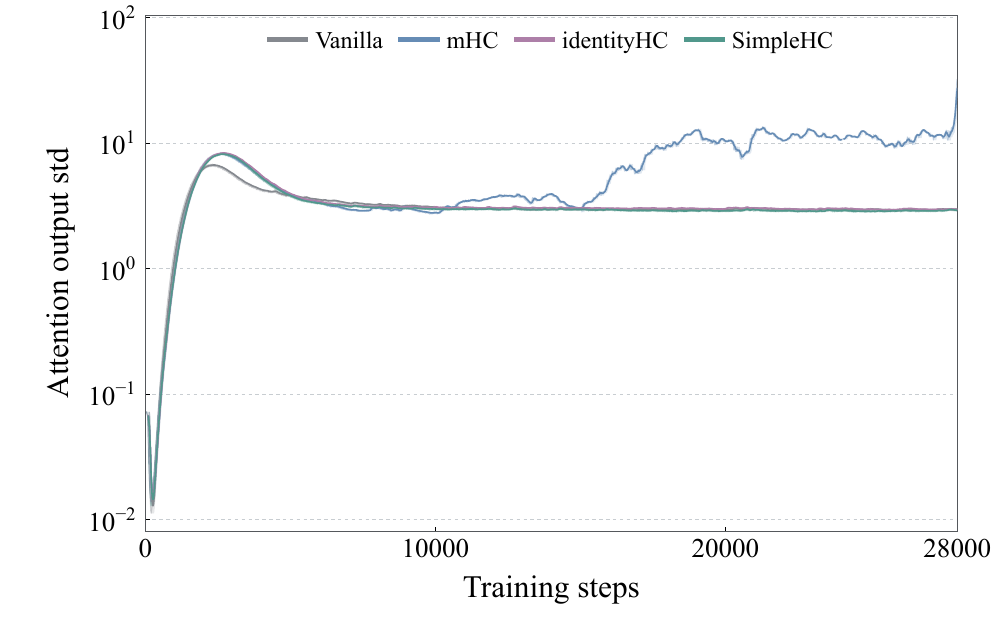}
\par
{\small (a) Attention output}
\end{minipage}\hfill
\begin{minipage}[t]{0.49\textwidth}
\centering
\includegraphics[width=\linewidth]{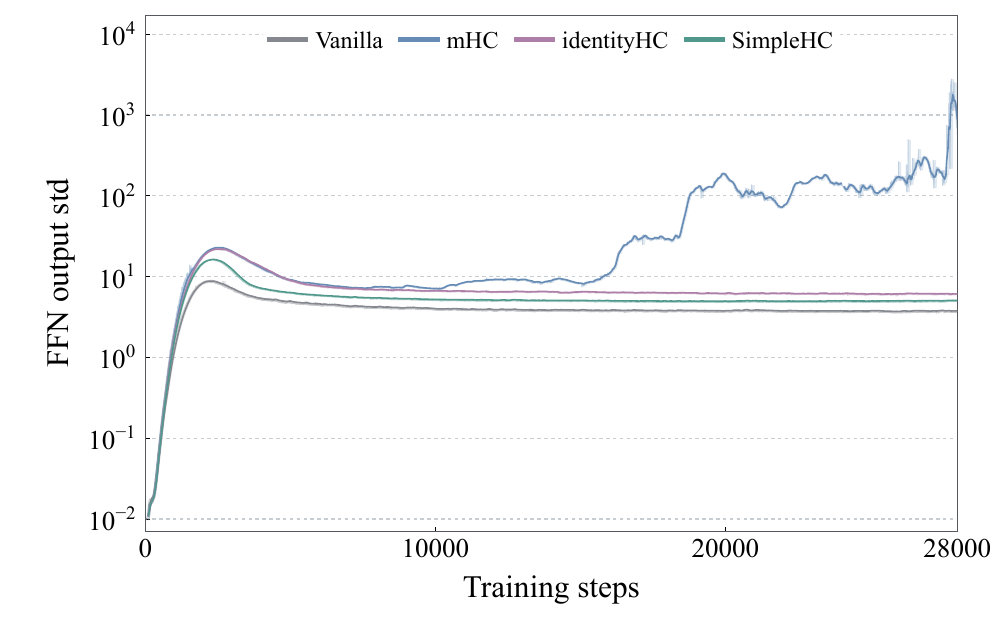}
\par
{\small (b) FFN output}
\end{minipage}

\caption{\textbf{Attention and FFN output standard deviations under stress.}
Lines show 100-step moving averages; shaded bands show the minimum
and maximum within each window.
Both panels use logarithmic vertical axes and cover the same 28,000 steps.}
\label{fig:stress-branch-outputs}

\end{figure}

\textbf{Stress testing.}
Under stress, mHC exhibits gradient spikes reaching approximately
$1.9\times10^6$, whereas SimpleHC and identityHC remain bounded (Figure~\ref{fig:training-stability}(b)). These spikes in mHC are
accompanied by loss spikes (Figure~\ref{fig:stress-loss}) and growing
branch-output scales, especially in the FFN. SimpleHC and identityHC
show no comparable sustained growth
(Figure~\ref{fig:stress-branch-outputs}), supporting stable training
despite the structural and execution simplifications.

\FloatBarrier
\subsection{Analysis of Stress-Test Results}
\label{sec:stability}
SimpleHC and identityHC maintain stable optimization under the combined
stress setting, indicating that removing residual mixing, with or without
shared gate inputs, can preserve training stability. SimpleHC remains
comparable to identityHC despite sharing gate inputs across sublayers.
In contrast, mHC exhibits instability in both forward activations and
backward gradients. To investigate this behavior, we inspect its branch
outputs, residual mixing matrices, and gradient paths.

In a captured early-layer MoE event, an input RMS of 0.294 produced a
branch-output RMS of $2.31\times10^4$ before write gating, locating the
initial large values in the FFN rather than residual mixing at this
sublayer. In the observed stress setting, 20 Sinkhorn iterations did not
always suffice to produce a doubly stochastic matrix. In one captured
case, row sums remained $(2,1,0.5,0.5)$, increasing same-token RMS from
$5.42\times10^4$ to $7.50\times10^4$. Mixing could therefore further
amplify existing FFN outliers. These observations also suggest a possible
rationale for the SwiGLU clamping used in DeepSeek-V4~\citep{xu2026deepseek}:
limiting FFN outliers before they enter the residual pathway. Residual mixing also introduces an additional gradient path because
$H^{\mathrm{res}}$ depends on the input. For $Y=H^{\mathrm{res}}(X)X$,
the differential is
$\mathrm{d}Y=H^{\mathrm{res}}\mathrm{d}X+(\mathrm{d}H^{\mathrm{res}})X$:
the first term propagates through the fixed matrix, while the second
captures changes in the mixing matrix itself. In one captured backward
spike, the gate-input gradient norm was approximately $9.4\times$ that
of the fixed-matrix path in the same call. Bounds computed from the
recorded read/write gates and upstream gradients made their contributions
negligible, implicating the dynamic mixing path. Whether this anomaly
reflects mathematical sensitivity or numerical effects in backward
remains unresolved. Identity transitions set
$\mathrm{d}H^{\mathrm{res}}=0$, eliminating this additional path,
consistent with the stable training observed in identityHC and SimpleHC.

\end{document}